\documentclass[10pt,twocolumn]{IEEEtran}
\IEEEoverridecommandlockouts
\usepackage{balance}
\usepackage{float}
\usepackage{cite}
\usepackage{amsmath,amssymb,amsfonts}
\usepackage{textcomp}
\usepackage{xcolor}
\usepackage{colortbl}
\usepackage{booktabs}
\usepackage{color}
\usepackage{amsmath}
\usepackage[linesnumbered,ruled,vlined]{algorithm2e}
\usepackage{hyperref}
\usepackage{cleveref}
\hypersetup{hidelinks}
\usepackage{multirow}
\usepackage{booktabs}
\usepackage{bm}
\usepackage[normalem]{ulem}

\usepackage{tikz}
\usepackage{pgfplots}
\usepackage{microtype}
\usepackage{graphicx}
\usepackage{adjustbox}
\usepackage{xspace}
\usepackage{thmtools, thm-restate}
\usepackage{pifont}
\usepackage{enumitem}

\usepackage{subcaption}
\usepackage{tikz}
\usetikzlibrary{arrows.meta, positioning, shapes.geometric}

\ifCLASSOPTIONcompsoc

\else

\fi

\def\BibTeX{{\rm B\kern-.05em{\sc i\kern-.025em b}\kern-.08em
    T\kern-.1667em\lower.7ex\hbox{E}\kern-.125emX}}

\newtheorem{remark}{Remark}

\newcommand{\blue}[1]{{{\color{blue} #1}}}

\NewDocumentCommand{\hjy}{ mO{} }{\textcolor{cyan}{\textsuperscript{\textit{hjy}}\textsf{\textbf{\small[#1]}}}}

\pgfplotsset{compat=1.18}

\definecolor{blue}{rgb}{0.0, 0.0, 0.0}

\begin{document}

\title{AdaptFly: Prompt-Guided Adaptation of Foundation Models for Low-Altitude UAV Networks}

\author{Jiao~Chen,
        Haoyi~Wang, Jianhua~Tang,~\IEEEmembership{Senior~Member,~IEEE},~and~Junyi~Wang,~\IEEEmembership{Member,~IEEE} 
\IEEEcompsocitemizethanks{
\IEEEcompsocthanksitem This research work has been submitted to IEEE for peer review and potential publication. Should the copyright be transferred, this version may become inaccessible without prior notice.
\IEEEcompsocthanksitem J. Chen, H. Wang, and J. Tang are with the Shien-Ming Wu School of Intelligent Engineering, South China University of Technology, Guangzhou 511442, China. J. Tang is also with the Key Laboratory of Cognitive Radio and Information Processing, Ministry of Education (Guilin University of Electronic Technology), Guilin, 541004, China. J. Wang is with the School of Information and Communication, Guilin University of Electronic Technology, Guilin 541004, China. The corresponding author is Jianhua Tang.
E-mails: \{202110190459, 202420160748\}@mail.scut.edu.cn, jtang4@e.ntu.edu.sg, wangjy@guet.edu.cn.
\IEEEcompsocthanksitem This work was supported in part by the National Key R\&D Program of China under Grant 2024YFE0200500, in part by the Guangdong Basic and Applied Basic Research Foundation under Grant 2024A1515012615, in part by the Department of Science and Technology of Guangdong Province under Grant 2021QN02X491, and in part by the Key Laboratory of Cognitive Radio and Information Processing, Ministry of Education under Grant CRKL240201.
}
}

\maketitle
\begin{abstract}

Low-altitude Unmanned Aerial Vehicle (UAV) networks rely on robust semantic segmentation as a foundational enabler for distributed sensing-communication-control co-design across heterogeneous agents within the network.
However, segmentation foundation models deteriorate quickly under weather, lighting, and viewpoint drift. Resource-limited UAVs cannot run gradient-based test-time adaptation, while resource-massive UAVs adapt independently, wasting shared experience.
To address these challenges, we propose AdaptFly, a prompt-guided test-time adaptation framework that adjusts segmentation models without weight updates. 
AdaptFly features two complementary adaptation modes. For resource-limited UAVs, it employs lightweight token-prompt retrieval from a shared global memory. For resource-massive UAVs, it uses gradient-free sparse visual prompt optimization via Covariance Matrix Adaptation Evolution Strategy.
An activation-statistic detector triggers adaptation, while cross-UAV knowledge pool consolidates prompt knowledge and enables fleet-wide collaboration with negligible bandwidth overhead. 
Extensive experiments on UAVid and VDD benchmarks, along with real-world UAV deployments under diverse weather conditions, demonstrate that AdaptFly significantly improves segmentation accuracy and robustness over static models and state-of-the-art TTA baselines.
The results highlight a practical path to resilient, communication-efficient perception in the emerging low-altitude economy.

\end{abstract}
\IEEEoverridecommandlockouts
\begin{IEEEkeywords}
UAV Networks, Foundation Model, Test-Time Adaptation, Semantic Segmentation, Low-altitude Economy
\end{IEEEkeywords}
\IEEEpeerreviewmaketitle

\section{Introduction}
Low-altitude UAV networks~\cite{10680081,11017717,10195219} are emerging as foundational infrastructure for the low-altitude economy. They provide ubiquitous communication and edge computing support to diverse applications such as remote sensing~\cite{11131292} and semantic communications~\cite{10007890,10778256}.
Modern UAVs are routinely deployed for urban traffic monitoring~\cite{10111056}, precision agriculture~\cite{9699056}, and disaster response, offering flexible, low-cost, and wide-area data acquisition.

A core enabler of low-altitude networked UAVs is semantic segmentation, the pixel-wise classification of aerial imagery that underpins fine-grained environmental perception and serves as a common semantic language for coordinated decision-making across UAV fleets~\cite{11131292,8951276,10756604,10599118}.
These environment features are fused with the radio environment map and processed by a deep neural network to predict coverage or signal strength.
Moreover, semantic segmentation serves as a bridge for network-level intelligence, enabling cross-domain extensions such as EdgeNet~\cite{8951276} for real-time road-scene understanding, SRNet~\cite{10756604} for spectral-domain signal parsing in 6G communications, and image-based predictive channel modeling~\cite{10294072,10599118} for propagation inference.
Recent surveys also highlight its potential in channel knowledge construction and environment-aware communications for 6G networks~\cite{10430216}.
Together, these advances promote robust sensing–communication–control co-design across distributed nodes~\cite{11045436}, strengthening the overall service quality and resilience of low-altitude UAV networks.

\begin{figure}[t]
    \centering
    \includegraphics[width=1\linewidth]{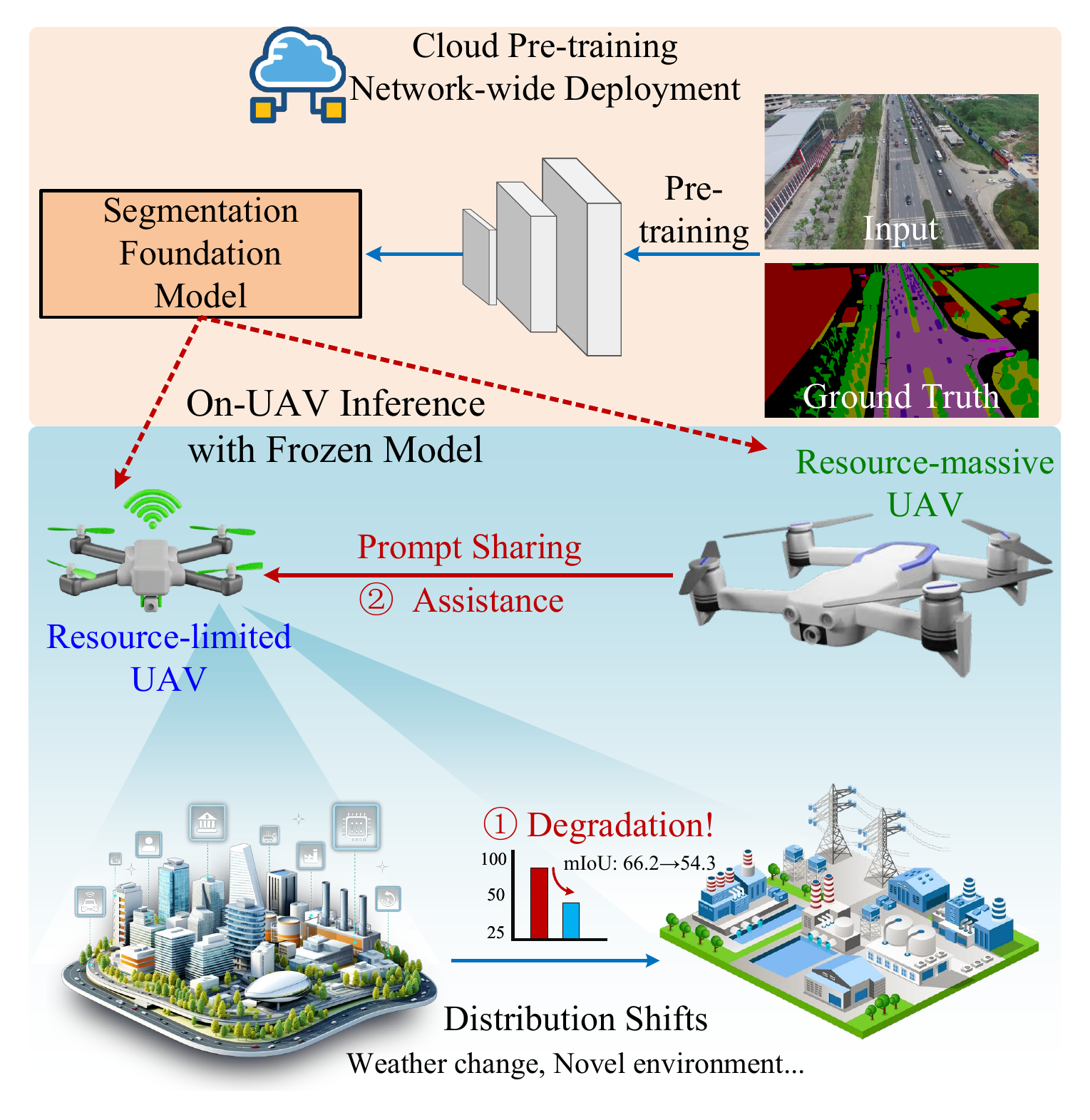}
    \caption{
    Heterogeneous UAV networks with collaborative adaptation.
    \ding{172} Both resource-limited and resource-massive UAVs face performance degradation.
    \ding{173} Collaborative assistance via shared prompts.
    }
    \label{fig:setting}
    \vspace{-0.1in}
\end{figure}

A major obstacle hindering reliable perception across UAV networks is distribution shift encountered when models deployed on multiple agents face diverse and dynamic environments.
In practice, the visual appearance of scenes can vary significantly over time and space. For example, changing lighting conditions (day vs. night, sunny vs. overcast), weather effects (fog, rain), seasonal and terrain differences, or sensor changes can all induce a shift in data distribution relative to the training domain~\cite{11169307,10436051}. 
Consequently, a segmentation model trained offline on one dataset often suffers degraded performance when deployed fleet-wide. This undermines the collective intelligence and coordination capabilities of the UAV network~\cite{wang2024unified,he2025satellite,cao2024computational,poorvi2025reliable}.
Fig.~\ref{fig:setting} illustrates a typical deployment scenario of heterogeneous UAVs facing domain shift and collaborative adaptation challenges.

Test-Time Adaptation (TTA) has emerged as a promising paradigm to tackle distribution shifts at the network edge, enabling models to adapt on the fly during inference without relying on labeled data or extensive retraining.
Recent advances have demonstrated that TTA can effectively improve model generalization under unseen domains using only unlabeled test data~\cite{wang2022continual}.
Representative approaches explore entropy minimization and continual adaptation strategies, achieving encouraging results in image classification tasks under domain shift~\cite{wangtent,niu2022efficient}.
However, applying TTA to dense prediction problems such as semantic segmentation remains challenging due to fine-grained spatial variations and limited robustness of normalization-based adaptation.

However, applying TTA to semantic segmentation is far more challenging. High-dimensional output (per-pixel predictions) and class imbalance in segmentation amplify the risk of erroneous updates. Indeed, recent studies found that naive adaptation strategies like batch norm tuning yield at best marginal gains and can even degrade segmentation performance in unseen domains. While more sophisticated schemes (e.g. teacher–student models or continual update rules) can improve stability over sequential frames, they often fail to consistently improve accuracy under complex, continuously shifting conditions. 
In summary, existing TTA methods struggle to deliver reliable semantic segmentation across UAV fleets: they lack mechanisms to leverage network-level synergies, while resource heterogeneity and real-time constraints further exacerbate the challenge.

Parallel to advances in test-time learning, prompt tuning~\cite{jia2022visual} has gained traction as an efficient paradigm for adapting large foundation models in both language and vision domains~\cite{our_comst}. Prompt tuning avoids full model retraining by inserting a small set of learnable parameters, known as prompts, into input or latent representations of models to guide frozen models to new tasks or domains.
Originally popularized in NLP, this approach has been adapted to vision transformers with notable success. 
For example, Visual Prompt Tuning (VPT)~\cite{jia2022visual} prepends a handful of learnable token vectors to input sequences of ViTs, achieving adaptation with less than 1\% additional parameters while keeping backbones frozen. 
Building on this idea, 
L2P~\cite{wang2022learning} proposes maintaining a prompt pool in memory and retrieving suitable prompts based on features of current inputs, enabling dynamic, context-aware model adjustment. Prompt-based tuning has even been explored in segmentation settings. For instance, APSeg~\cite{he2024apseg} generates visual prompts through meta-learning and uses a query-focused attention mechanism to achieve few-shot segmentation on new domains. These works demonstrate that prompt tuning can serve as a lightweight adaptation mechanism for vision models. 

However, most prompt tuning methods operate in an offline or episodic training regimes. These methods typically assume access to target-domain training samples or class labels and often optimize prompts via gradient descent during fine-tuning stages. In real-time test-time scenarios of UAV deployments, we cannot afford iterative gradient updates or any form of supervised re-training. Prompt tuning in this context must be gradient-free, fast, and resource-aware, which current methods do not address. Moreover, prior studies typically consider single models adapting in isolation; they do not exploit the potential of networked UAV fleets to share learned prompt knowledge, an aspect that could greatly enhance robustness in practice.

In light of these gaps, we propose \textbf{AdaptFly}, a unified prompt-guided TTA framework for networked heterogeneous UAV networks in dynamic environments. AdaptFly dynamically adjusts segmentation models at inference time using visual prompts, without any gradient-based updates to model weights. It is designed to be aware of the computational capabilities of each UAV and to facilitate cross-UAV knowledge transfer through shared memory, thereby aligning with the cognitive communications vision of autonomous and collaborative learning in networked UAVs.
The main contributions of this work are summarized as follows:

$\bullet$ Network-centric framework: AdaptFly enables fleet-wide adaptive perception by unifying prompt-based adaptation with asynchronous knowledge consolidation, transforming isolated UAV agents into a collaborative network that collectively improves resilience under distribution shifts.

$\bullet$ Network-assisted prompt retrieval for resouce-limited UAV: For low-power UAVs, AdaptFly enables efficient, training-free adaptation via prompt retrieval from a global memory. Each UAV selects suitable token prompts based on context embedding, allowing fast model adjustment without any on-device optimization.

$\bullet$ Prompt optimization for resouce-massive UAV: For UAVs with sufficient compute capacity, AdaptFly employs a black-box optimization strategy based on Covariance Matrix Adaptation Evolution Strategy (CMA-ES) to adapt sparse visual prompts (SVP) at test time. This evolutionary tuning focuses on high-uncertainty regions and avoids back-propagation, enabling efficient adaptation while preserving the frozen backbone.

$\bullet$ Fleet-wide knowledge consolidation: We design an activation-statistics-based domain drift detector to decide when adaptation is needed. Adapted prompts are consolidated into a global memory and made available to other UAVs, enabling asynchronous, communication-efficient knowledge sharing across the fleet.

$\bullet$ Real-world validation on UAV testbed: We implement and evaluate AdaptFly on a physical UAV platform under dynamic scenes such as dusk and rain. The results demonstrate its effectiveness and efficiency in realistic low-altitude environments, supporting practical deployment in the emerging low-altitude economy.

\section{Related Works}
\subsection{Semantic Segmentation in Low-Altitude UAV Networks}

\begin{figure}[t]
    \centering
    \includegraphics[width=1\linewidth]{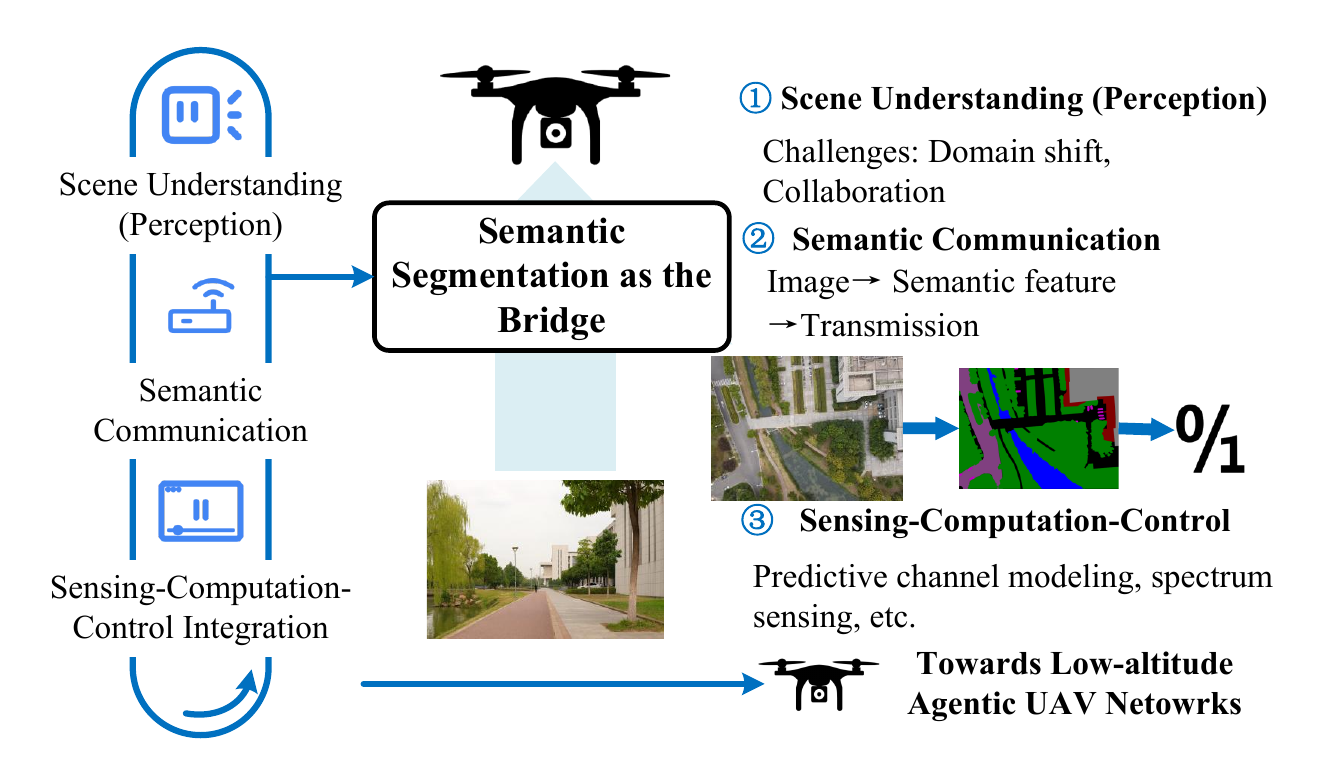}
    \caption{
    Semantic segmentation as the perception backbone of low-altitude UAV networks.
    }
    \label{fig:as the perception}
    \vspace{-0.1in}
\end{figure}

In low-altitude UAV networks, semantic segmentation serves not merely as a perception module but as a pivotal enabler that bridges sensing, communication, computation, and control~\cite{wang2025uavscenes,sunderraman2024uav3d,11131292,11017717}. 
Aerial imagery captured in urban monitoring~\cite{10111056}, precision agriculture~\cite{9699056}, and post-disaster response exhibits oblique viewpoints, ultra-high resolution, and significant scale variations, posing unique challenges for pixel-wise classification~\cite{10638762,10736665,9875022,9749130}.

Recent advances address these challenges by tailoring segmentation to constraints and opportunities of low-altitude operations~\cite{xue2025cacp,10498080}. For instance, DroneSegNet~\cite{9689963} integrates elevation maps with RGB images and employs a bidirectional ConvLSTM to model spatial-height semantic relationships, significantly enhancing robustness in complex urban environments. 
To combat shortcut bias induced by scene deformation in disaster zones, BCU-Net proposes a bias-compensation augmentation learning framework coupled with a reinforcement learning-based computation offloading strategy, achieving higher segmentation accuracy while reducing processing latency~\cite{10476687}.

Crucially, semantic segmentation is evolving from an isolated perception task into a cornerstone of intelligent coordination within UAV networks~\cite{wang2025uavscenes,sunderraman2024uav3d}.
On one hand, high-quality segmentation guides service-oriented perception and enables cross-domain intelligence. It supports real-time road-scene understanding~\cite{8951276}, spectral-domain signal parsing in 6G communications~\cite{10756604}, and image-based predictive channel modeling~\cite{10294072,10599118}. It also improves the efficiency of mobile edge computing (MEC) tasks.
On the other hand, segmentation outputs serve as input sources for channel knowledge construction~\cite{10430216} semantic communication. By transforming modalities (e.g., image-to-text), they drastically reduce transmission overhead. When combined with UAV collaborative relaying, they jointly optimize quality of experience and energy consumption~\cite{10930657}.

Moreover, in adversarial environments, semantic segmentation enables anti-jamming communication by transmitting only essential semantic features instead of raw data, reducing vulnerability to interference and improving task efficiency~\cite{10678860}. This evolution positions semantic segmentation as a foundational bridge linking perception, communication, computation, and control, enabling efficient, privacy-preserving, and resilient low-altitude UAV networks.

\subsection{Prompt Tuning for Vision Foundation Models}

Prompt tuning has emerged as a parameter-efficient adaptation technique that enables large foundation models to generalize to new tasks or domains without full fine-tuning~\cite{jia2022visual}. By inserting a small number of learnable prompt vectors into model inputs, it enables fast adaptation to new tasks or domains while keeping backbone parameters frozen. 

VPT~\cite{jia2022visual} introduces learnable prompts before patch tokens in ViT~\cite{dosovitskiy2020image}, requiring only minimal parameter updates. 
L2P~\cite{wang2022learning} proposes a prompt pool with retrieval based on task embeddings, offering dynamic and context-aware adaptation. 
ApSeg~\cite{he2024apseg} integrates meta prompt generation and query-attention routing to achieve few-shot segmentation in vision tasks.

Recent advances also explore deployment of prompt tuning in mobile edge networks with constrained resources~\cite{10891165}. For instance, a unified framework for guiding generative AI via wireless perception in edge-constrained settings~\cite{wang2024unified}, motivates test-time lightweight adaptation in UAV scenarios.
However, most existing methods focus on offline training or continual learning, and they often assume access to task labels or fine-tuning stages. They are not designed for test-time scenarios requiring gradient-free, online, and resource-aware adaptation.

\subsection{Test-Time Adaptation}

TTA aims to improve model generalization under distribution shift in online inference settings, without accessing source data~\cite{wang2022continual}. Classical methods such as TENT (entropy minimization)~\cite{wangtent} and EATA~\cite{niu2022efficient} (Fisher-based regularization) optimize partial model parameters by minimizing unsupervised objectives such as prediction entropy or pseudo-label reconstruction. CoTTA~\cite{wang2022continual} further improves robustness by introducing temporal consistency and perturbation-based augmentation.

Although these methods perform well in image classification, they encounter significant challenges in semantic segmentation. Recent studies have shown that normalization-based strategies (e.g., TENT~\cite{wangtent}) offer limited improvements and may even degrade performance in segmentation tasks~\cite{yi2024question}. Moreover, while Teacher-Student approaches help stabilize adaptation under noisy pseudo-labels and temporal correlation, they often fail to deliver consistent gains under complex or continuously shifting distributions.

Recent advances in embodied intelligence for low-altitude economies emphasize the need for integrated adaptation mechanisms. An ISC3 paradigm that jointly considers perception, communication, computation, and control~\cite{yang2024embodied} aligning well with goals of robust TTA under dynamic UAV environments.

\section{Problem Statement and Motivation}
\subsection{Problem Definition}

Before formally defining the problem, we first illustrate in Fig.~\ref{fig:as the perception} how semantic segmentation functions as the perception backbone that bridges sensing, communication, computation, and control in low-altitude UAV networks. This figure provides a conceptual overview of the research problem and motivates the need for adaptive segmentation across heterogeneous UAVs.

Building on this overview, we now formalize the problem setting for adaptive semantic segmentation in networked UAVs.
Let \( f_\theta(\cdot) \) be a semantic segmentation foundation model trained on a labeled source-domain dataset \( \mathcal{D}_{\text{train}} = \{(\mathbf{x}_i, y_i)\} \), where \( \mathbf{x}_i \sim P(\mathbf{x}) \). 
In the deployment phase, this model is distributed to multiple low-altitude UAVs forming a networked fleet, where each UAV performs image perception tasks while collectively adapting to evolving environmental conditions.

However, in real-world scenarios, test images often come from environments with distributions different from the training domain, denoted as \( Q(\mathbf{x}) \), where \( Q(\mathbf{x}) \ne P(\mathbf{x}) \). This reflects domain shift issues caused by viewpoint changes, lighting variations, or structural layout differences, significant drops in model performance when facing unseen domains.

\begin{figure*}[!ht]
    \centering
    \includegraphics[width=0.7\linewidth]{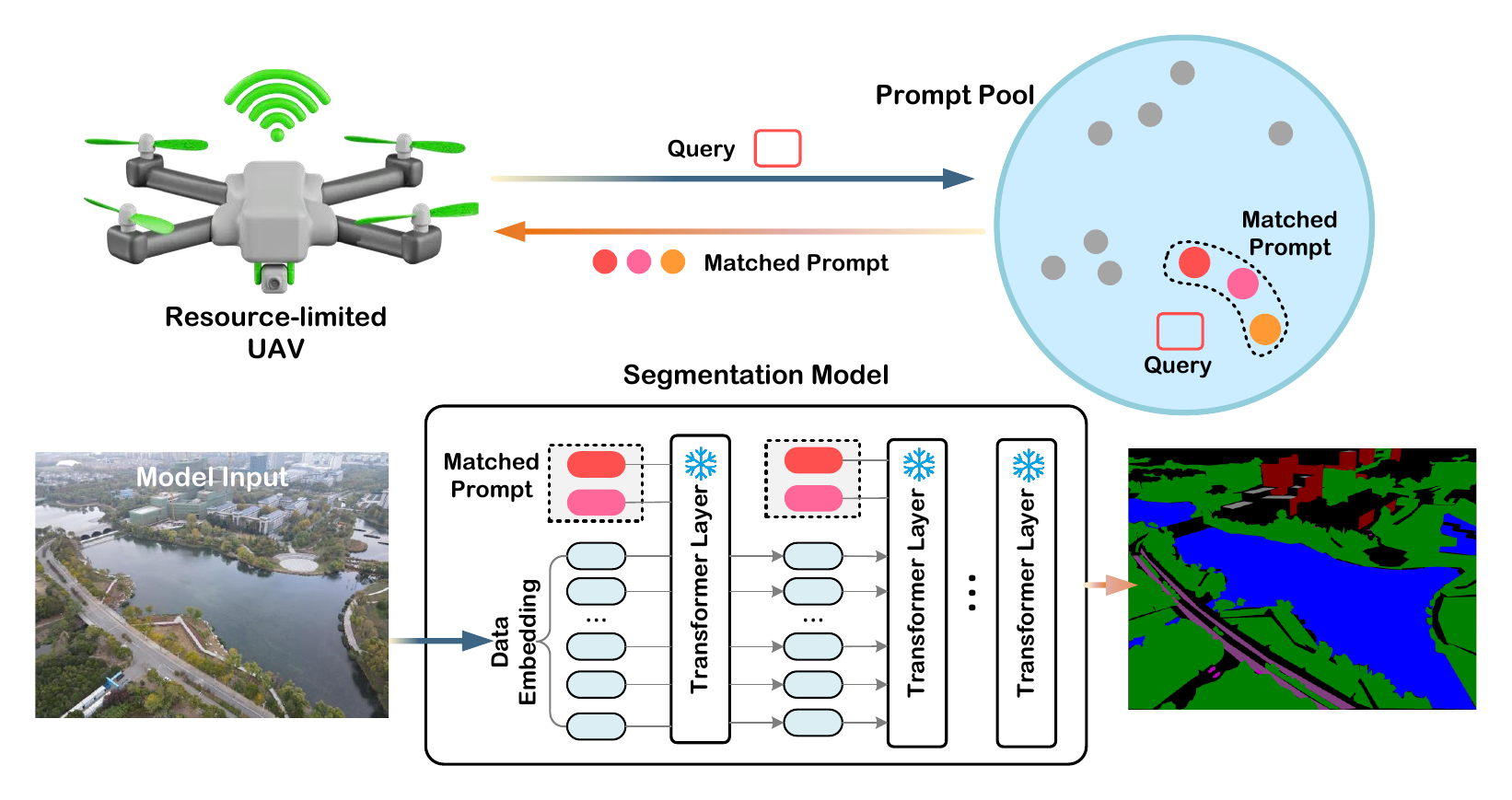}
    \caption{Token prompt retrieval pipeline for resource-limited UAVs.}
    \label{fig:token-retrieval}
    \vspace{-0.1in}
\end{figure*}

To enhance model adaptability to such shifts during inference, a commonly used approach is TTA. Under TTA, backbone parameters \( \theta_f \) are frozen, and only a small number of learnable parameters \( \theta_l \) are optimized by minimizing a self-supervised or unsupervised loss:

\begin{equation}
    \min_{\theta_l} \mathcal{L}(\mathbf{x}; \theta_f, \theta_l), \quad \mathbf{x} \sim Q(\mathbf{x}),
\end{equation}
where \( \mathcal{L} \) could be entropy minimization, pseudo-label loss, consistency regularization~\cite{wang2022continual}, etc.

\subsection{Motivation}

\blue{We define resource-limited UAVs as onboard systems with less than 10 TOPS computational capacity (e.g., Jetson Nano), which are commonly deployed for mid-complexity tasks such as aerial navigation, mapping, and real-time object detection.
Conversely, resource-massive UAVs refer to platforms equipped with 30 TOPS or higher computing power (e.g., Jetson Orin NX), suited for high-performance applications such as autonomous driving, intelligent surveillance, or complex edge-AI inference.}

Although TTA enables model self-adjustment in unseen domains, most existing methods follow the paradigm of independent adaptation on each device, which leads to inefficiencies in multi-UAV scenarios due to the following problems:

\begin{itemize}
  \item Redundant learning overhead: Multiple UAVs may encounter similar scenes at different times, but all start adapting from scratch, wasting computational resources.
  \item Lack of knowledge sharing: Adaptation knowledge cannot be transferred across devices, resulting in poor system-wide efficiency.
  \item Resource limitations: Resource-limited UAV platforms lack capacity to perform real-time adaptation via gradient-based optimization.
  \item Communication constraints: Transmitting raw images or full model parameters is costly and may not be feasible in distributed settings.
\end{itemize}

In networked low-altitude operations, UAVs often share overlapping coverage areas and encounter correlated environmental conditions due to spatial proximity and synchronized mission timelines.
This spatial-temporal correlation creates an opportunity: adaptation knowledge acquired by one UAV can benefit others, reducing redundant learning and improving fleet-wide resilience. However, existing TTA methods lack mechanisms to exploit such network-level synergies, motivating our cross-UAV knowledge sharing design.

\subsection{Prompt Representations}
To address the above challenges, we introduce prompt modules as external, lightweight adaptation components. These prompts serve as dynamic inputs to steer model behavior without modifying backbones. They can be unified under a general low-rank perturbation formulation:
\begin{equation}
    x' = x \oplus G(\mathbf{p}),
\end{equation}
where \(x\) is an input image (or an intermediate feature),  
\(\mathbf{p}\) is a set of prompt parameters, and \(G(\cdot)\) projects them to the proper domain. \blue{$\oplus$ denotes composition: concatenation for token prompts~\eqref{eq:token_prompts} 
and pixel-wise addition for SVP~\eqref{eq:svp}.}
We consider two specific realizations of this general formulation: \textbf{token prompts} and \textbf{SVP}.

\paragraph{Token Prompts}
Token prompts~\cite{jia2022visual} are learnable vectors inserted into input sequence of the transformer. Given an embedded patch sequence \(\mathbf{Z}_0 \in \mathbb{R}^{N\times d}\) (\(N\) patches, hidden size \(d\)), we introduce an additional learnable matrix
\(
\mathbf{P}\!=\![\mathbf{p}_1;\dots;\mathbf{p}_L] \in \mathbb{R}^{L\times d},
\)
called \emph{token prompts}:

\begin{equation}\label{eq:token_prompts}
    \widetilde{\mathbf{Z}}_0=[\mathbf{P};\;\mathbf{Z}_0] \in \mathbb{R}^{(L+N)\times d},
\end{equation}
which is fed to all subsequent transformer layers without modifying backbone weights.  
Typical hyper-parameters in segmentation backbones are  
\(L\in\{4,8,16\}\) tokens and \(d=768\) (e.g., SegFormer~\cite{xie2021segformer}), 
yielding only \(L\!\times\!d\approx(3\text{--}12)\times10^{3}\) additional parameters.
During adaptation, \(\mathbf{P}\) is optimized while backbones remains frozen.

\paragraph{Sparse Visual Prompts}
SVP~\cite{yang2024exploring} act directly in the image space by perturbing a small subset of pixels.  
Given an input image \(\mathbf{x}\!\in\!\mathbb{R}^{3\times H\times W}\), we create a sparse perturbation map \(\mathbf{p}\) with

\begin{equation}\label{eq:svp}
    \| \mathbf{p} \|_{0}=K,\quad K = \rho HW,\quad
\tilde{\mathbf{x}} = \mathbf{x} + \mathbf{p},
\end{equation}
where \(\rho\) is the sparsity ratio.  
Following prior work, we set \(\rho=10^{-3}\) (i.e., \ top \(0.1\%\) uncertain pixels selected via MC-Dropout).  
Only RGB values at those \(K\) locations are learnable (\(3K\) parameters), giving a perturbation budget comparable to token prompts yet fully aligned with pixel semantics.  
Pixel locations are refreshed online at every time step, the perturbation values are updated by a self-supervised consistency loss while backbones stays fixed.

Both prompt types instantiate the same low-rank formulation but differ in injection domain: token prompts operate in feature space, while SVP act in image space. In the following section, we develop resource-aware adaptation strategies based on these two forms.

\section{Prompt-Guided Adaptation for Heterogeneous UAV Networks}
To cope with dynamic low-altitude environments faced by networked UAV fleets, we propose \textbf{AdaptFly}, a unified prompt-guided framework for collaborative and resource-aware test-time adaptation across heterogeneous UAV agents.
AdaptFly operates under a frozen backbone (e.g., SegFormer~\cite{xie2021segformer}) and achieves domain adaptation by injecting lightweight prompts. Importantly, it accommodates heterogeneous UAV agents by adopting resource-aware strategies tailored to both low-end and high-end computational settings.

\subsection{Resource-Limited UAVs: Network-Assisted Token Prompt Retrieval}
\label{sec:uav_l}
For resource-limited UAVs, AdaptFly employs a forward-only, training-free adaptation scheme based on token prompt retrieval. Fig.~\ref{fig:token-retrieval} provides a workflow example of this process.
Inspired by memory-based architectures of L2P~\cite{wang2022learning} and UAV task offloading~\cite{10736665}, we maintain a key–value prompt pool:
\begin{equation}
    \mathcal{M}=\bigl\{(\mathbf{k}_i,\mathbf{P}_i)\bigr\}_{i=1}^{K},
\end{equation}
where each key $\mathbf{k}_i\!\in\!\mathbb{R}^{d}$ is learnable, and each associated value $\mathbf{P}_i\!\in\!\mathbb{R}^{L\times d}$ consists of $L$ token vectors compatible with the frozen SegFormer backbone. This prompt pool $\mathcal{M}$ is either pre-trained offline or dynamically enriched by knowledge consolidated from resource-massive UAVs.

\textbf{Instance-wise query.}
Given a test image $\mathbf{x}$, a query embedding $\mathbf{q} = g(\mathbf{x})$ is computed using a frozen image encoder.
Similarity between $\mathbf{q}$ and each key $\mathbf{k}_i$ is measured using cosine distance. The top-$N$ most relevant keys are selected as:
\begin{equation}
\mathcal{I} = \text{TopN}(\mathbf{q}, \mathcal{M}).
\end{equation}

\textbf{Prompt assembly.}  
Corresponding prompts are concatenated in descending similarity order,
\begin{equation}
    \mathbf{P}_{\text{sel}}=[\,\mathbf{P}_{i_1};\dots;\mathbf{P}_{i_N}\,]\in\mathbb{R}^{(NL)\times d},
\end{equation}
then prepended to the patch sequence:
\begin{equation}
    \widetilde{\mathbf{Z}}_0=[\,\mathbf{P}_{\text{sel}};\;\mathbf{Z}_0\,].
\end{equation}

The augmented sequence $\widetilde{\mathbf{Z}}_0$ is then passed through the transformer without back-propagation. This forward-only pipeline enables per-instance adaptation by dynamically reusing task-relevant prompts from memory. Owing to its lightweight nature and minimal computational cost, this approach is particularly well-suited for real-time deployment on resource-limited UAV agents.

\begin{figure*}[htbp]
\centering
\hspace{0.04\textwidth} 
\begin{subfigure}[t]{0.45\textwidth}
\centering
\includegraphics[height=7.0cm]{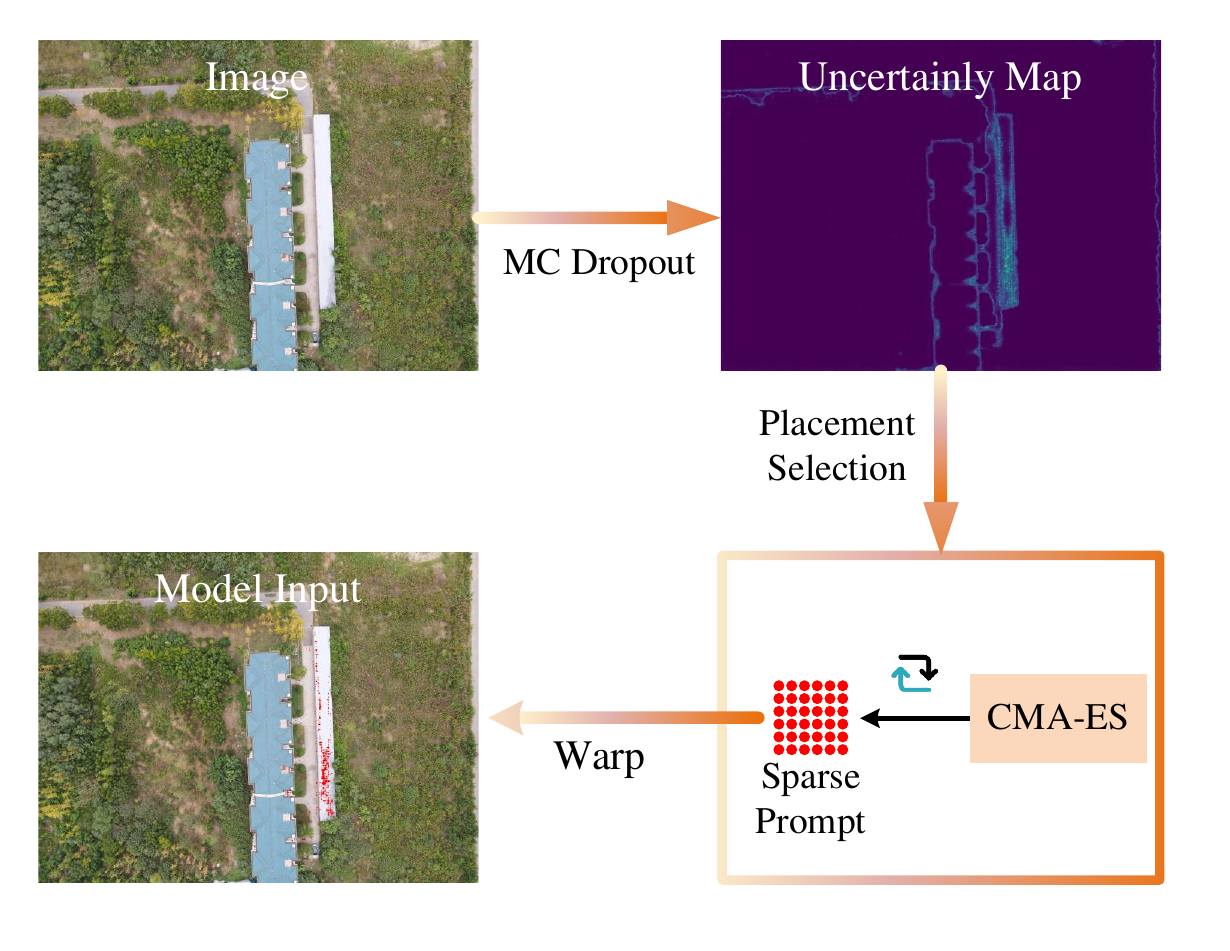}
\caption{SVP placement and model input}
\label{fig:pdf-subfig}
\end{subfigure}
\hfill
\begin{subfigure}[t]{0.48\textwidth}
\centering
\resizebox{!}{7.0cm}{ 
\begin{tikzpicture}[
    font=\large,
    block/.style={rectangle, draw=blue!60, fill=blue!10,
                  rounded corners, align=center,
                  minimum width=4.6cm, minimum height=1.0cm},
    decision/.style={diamond, draw=orange!70, fill=orange!10,
                     align=center, inner sep=1pt, aspect=2},
    arrow/.style={-Latex, thick}
]
\node[block]      (init)    at (0,5)   {Init $\mu^{(0)},\Sigma^{(0)}$};
\node[block, below=0.6cm of init] (samp)   {Sample $\mathbf{p}^{(j)} \sim \mathcal{N}(\mu,\Sigma)$};
\node[block, below=0.6cm of samp] (fit)    {Compute fitness~Eq.~\eqref{eq:cma-fitness}};
\node[block, below=0.6cm of fit]  (update) {Update $\mu,\Sigma$};
\node[decision, below=0.6cm of update] (stop) {Converged or\\ $t \ge T$?};
\node[block, below=0.6cm of stop] (out)    {Output $\mathbf{p}^{\star}$};

\draw[arrow] (init)   -- (samp);
\draw[arrow] (samp)   -- (fit);
\draw[arrow] (fit)    -- (update);
\draw[arrow] (update) -- (stop);
\draw[arrow] (stop)   -- node[right]{yes} (out);
\draw[arrow] (stop.west) -- ++(-1.5,0) |- node[left]{no} (samp.west);

\end{tikzpicture}
} 
\caption{CMA-ES based SVP optimization}
\label{fig:tikz-subfig}
\end{subfigure}
\caption{CMA-ES driven online SVP optimization.}
\label{fig:combined-flow}
\vspace{-0.1in}
\end{figure*}
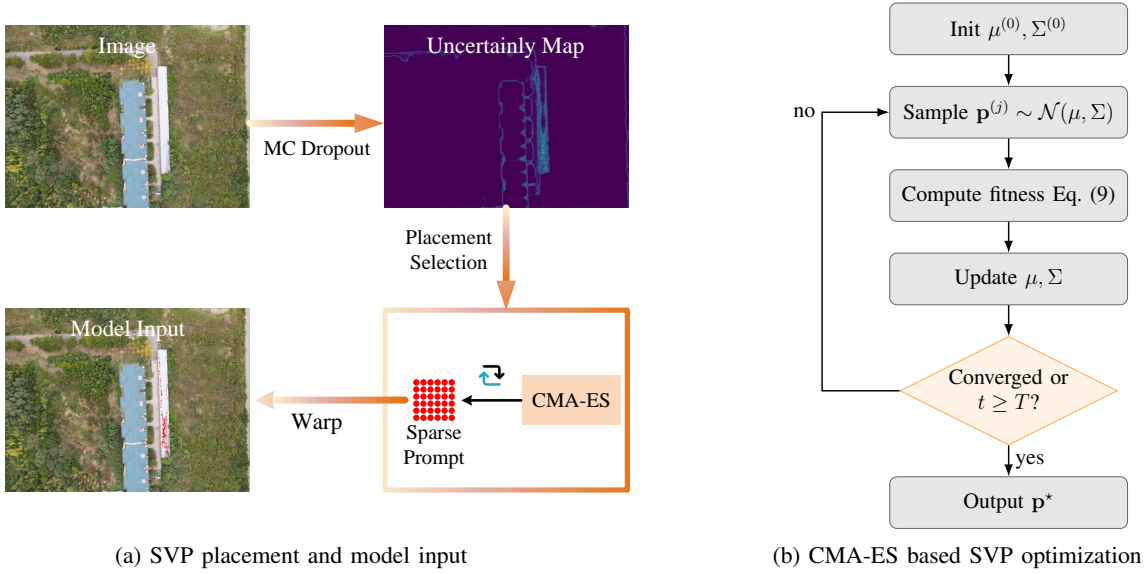

\subsection{Resource-Massive UAV: Sparse Prompt Optimization via CMA-ES}
\label{sec:uav_h}

When a UAV is equipped with a powerful onboard GPU or maintains a high-bandwidth
link to the cloud, it can afford on-the-fly prompt optimization rather than
mere prompt retrieval.
For these resource-massive UAVs, we adopt an enhanced SVP optimization strategy that utilizes a gradient-free evolutionary algorithm, CMA-ES~\cite{hansen2016cma}, to adapt prompts. Fig.~\ref{fig:combined-flow}(b) gives an overview.

Step\,1: Uncertainty-aware placement. 
We first run the frozen backbone
$f_{\theta}$ for $K$ times with Monte-Carlo dropout to obtain an uncertainty map
$\mathbf{U}\!\in\!\mathbb{R}^{H\times W}$.
We then select the $\rho HW$ most uncertain pixels and allocate one learnable prompt parameter per selected pixel. The resulting sparse mask is denoted by
$\mathbf{M}\!\in\!\{0,1\}^{H\times W}$.

Step\,2: CMA-ES search.
Let $\mathbf{p}\!\in\!\mathbb{R}^{3\times H\times W}$ be the prompt, where
$\mathbf{p}_{wh}\!=\!\mathbf{0}$ if $\mathbf{M}_{wh}=0$.
We seek the prompt that minimizes prediction uncertainty:
\begin{equation}
\label{eq:cma-fitness}
\mathcal{F}(\mathbf{p}) \;=\;
\mathcal{H}\!\bigl(f_{\theta}(\mathbf{x}\!+\!\mathbf{p})\bigr),
\end{equation}
where $\mathcal{H}$ denotes mean pixel-wise Shannon entropy.
Since the objective is non-differentiable with respect to the discrete sparsity mask, 
we employ CMA-ES~\cite{hansen2016cma} as a gradient-free optimizer. The full optimization procedure for a single image (executed when a significant domain shift is detected) is outlined in Algorithm~\ref{alg:svp_cma-es}, with an illustrative overview provided in Fig.~\ref{fig:combined-flow}(b).

\begin{algorithm}[!t]
\DontPrintSemicolon
\caption{SVP optimization with CMA-ES}
\label{alg:svp_cma-es}
\KwIn{Frozen model $f_{\theta}$, input image $\mathbf{x}$, sparsity mask $\mathbf{M}$,  
        population size $m$, elite size $m_e$, generations $T$, initial std.\ $\sigma$}
\KwOut{Optimised sparse prompt $\mathbf{p}^{\star}$}

\textbf{Init:} mean $\boldsymbol{\mu}\!\leftarrow\!\mathbf{0}$, covariance $\Sigma\!\leftarrow\!\sigma^{2}\mathbf{I}$\;  
\For{$t = 1$ \KwTo $T$}{                                                                       
  \For{$j = 1$ \KwTo $m$ \KwSty{in parallel}}{
      sample $\mathbf{p}^{(j)} \sim \mathcal{N}\!\bigl(\boldsymbol{\mu}, \Sigma\bigr)$\;
      $\mathbf{p}^{(j)} \leftarrow \mathbf{p}^{(j)} \odot \mathbf{M}$\; 
      $F^{(j)} \leftarrow \mathcal{H}\!\bigl(f_{\theta}(\mathbf{x}+\mathbf{p}^{(j)})\bigr)$    
  }
  $\mathcal{E} \leftarrow$ indices of the $m_e$ prompts with lowest $F^{(j)}$\;               
  update $(\boldsymbol{\mu},\Sigma)$ using $\{\mathbf{p}^{(j)}\}_{j\in\mathcal{E}}$\;            
}
\Return{$\displaystyle\mathbf{p}^{\star} = \arg\!\min_{j}\,F^{(j)}$}\;                         
\end{algorithm}

Step\,3: Prompt warping and inference.
Because UAV cameras may yaw or pitch between frames~\cite{vincent2025high}, 
we propagate the optimized prompt $\mathbf{p}^{\star}$ to the next frame via a lightweight spatial warp (Fig.~\ref{fig:combined-flow}(a), ``Warp''). This aligns the previous-frame embedding to the current image before feeding it into the model input. The prompt is refreshed only when the uncertainty map changes significantly. The amortized cost is therefore \(\mathcal{O}(1)\) per frame after
the initial CMA-ES run.

\begin{remark}
\noindent\textbf{Why CMA-ES?}\;
Back-propagation through high-resolution images requires substantial GPU
memory and may expose the model to adversarial gradients.
CMA-ES is derivative-free, stable under noisy fitness evaluations, and
converges in $\sim$30 generations for our $1024{\times}1024$ UAVid frames,
taking $<0.25$\,s on an RTX 4090 (batch 1).
\end{remark}

Overall, this SVP $+$ CMA-ES module offers a strong complement to the
token-prompt pipeline of Sec.~\ref{sec:uav_l}, enabling AdaptFly to exploit
the full computational envelope of heterogeneous UAV fleets.

\subsection{Domain Shift Detection via Activation Statistics for Prompt Management}

To enable dynamic adaptation under diverse testing conditions, AdaptFly incorporates a lightweight domain shift detection mechanism to determine whether input samples are drawn from previously unseen domains. This mechanism serves as a critical trigger for prompt retrieval (on resource-limited UAVs) and prompt optimization (on resource-massive UAVs), while also guiding maintenance and updates of the global prompt pool.

Specifically, we adopt an unsupervised shift detection method based on activation statistics from early layers of the segmentation backbone. This approach does not require access to source data or annotations and is well-suited for online deployment scenarios.

Let $f_\theta(\cdot)$ denote the backbone network (e.g., SegFormer~\cite{xie2021segformer}), and $f_\theta^s(\cdot)$ denote its stem layer. For each test input $\mathbf{x}_t$, we extract output features from $f_\theta^s(\mathbf{x}_t)$ and compute activation statistics as a tuple of mean and standard deviation:
\begin{equation}
    \hat{\phi}_t = (\hat{\mu}_t, \hat{\sigma}_t).
\end{equation}

A widely-used exponential moving average (EMA) of historical statistics is maintained as $\phi_d$, updated via exponential smoothing:
\begin{equation}
    \phi_d \leftarrow \lambda \hat{\phi}_t + (1 - \lambda)\phi_d,
\end{equation}
where $\lambda \in [0, 1]$ is a smoothing factor, and $\phi_d$ represents the estimated domain-specific distribution for the current UAV.

To determine whether a distributional shift has occurred, we compute symmetric KL-based divergence between current and historical statistics:
\begin{equation}
    D(\phi_d, \hat{\phi}_t) = \frac{1}{H} \sum_{i=1}^{H} \mathrm{KL}(\phi_{d,i} \parallel \hat{\phi}_{t,i}) + \mathrm{KL}(\hat{\phi}_{t,i} \parallel \phi_{d,i}),
\end{equation}
where $H$ is the dimensionality of statistics, and the KL divergence between two univariate Gaussians is defined as:
\begin{equation}
    \mathrm{KL}(\phi_1 \parallel \phi_2) = \frac{1}{2\sigma_2^2} \left( \sigma_1^2 + (\mu_1 - \mu_2)^2 \right),
\end{equation}
\blue{where $\mu_1, \mu_2$ and $\sigma_1, \sigma_2$ denote the mean and standard deviation
of activation features in the corresponding domains, respectively.
}
If the divergence exceeds a predefined threshold $z$, i.e., $D(\phi_d, \hat{\phi}_t) > z$, we determine that a domain shift has occurred.

\subsection{Cross-UAV Knowledge Consolidation and Sharing}
\label{subsec:convert}
AdaptFly implements a network-level knowledge consolidation mechanism that transforms locally optimized prompts into reusable assets, enabling fleet-wide generalization without redundant computation.
When a resource-massive $\mathcal{U}_h$ re-optimizes an SVP $\mathbf{p}^H$ via CMA-ES for a newly encountered domain (i.e., $D(\phi_d^H, \hat{\phi}_t^H) > z$), the acquired adaptation knowledge is made reusable by other UAVs.

To ensure compatibility with the token prompt pool $\mathcal{M}$ used by resource-limited $\mathcal{U}_\ell$ agents, we introduce a cross-format consolidation process. Specifically, the optimized SVP $\mathbf{p}^H$ is converted into a token prompt $\mathbf{P}_{\text{new}}$ by minimizing representation discrepancy:
\begin{equation}\label{eq:convert}
    \mathbf{P}_{\text{new}} = \arg\min_{\mathbf{P}} \left\| f_\theta([\mathbf{P}; \mathbf{Z}_0]) - f_\theta(\mathbf{x}_t^H + \mathbf{p}^H) \right\|_2^2,
\end{equation}
where $f_\theta(\cdot)$ denotes the frozen encoder, whose output is an encoder feature map. $Z_0$ is the default prompt context. 
\blue{Specifically, $Z_0$ represents the original token sequence 
used before adaptation, serving as the initialization of the prompt embedding.}
This ensures that both sides of Eq.~(\ref{eq:convert})—the concatenated token prompt $[\mathbf{P}; \mathbf{Z}_0]$ and the SVP-modulated input $\mathbf{x}_t^H + \mathbf{p}^H$—are represented in the same token-feature space, allowing direct feature-level alignment.
This offline distillation step requires only a few gradient steps and does not affect online test-time efficiency of $\mathcal{U}_\ell$.

Alternatively, when distillation is deferred, $\mathcal{U}_h$ may instead register a key-value pair $(\mathbf{q}^H, \text{domain id})$ in memory $\mathcal{M}$, with $\mathbf{q}^H = g(\mathbf{x}_t^H)$ representing the adapted domain query. Token prompts can be generated on demand when the same domain reoccurs.
This mechanism enriches the global prompt pool with real-time adaptation experiences, enabling low-resource UAVs to retrieve effective token prompts in similar environments, thereby improving system-wide adaptation efficiency and resilience.

\blue{To handle synchronization across geographically distributed UAVs, 
we adopt an \textit{asynchronous grow-and-refine strategy}. 
Each resource-massive $\mathcal{U}_h$ independently uploads its newly distilled prompts 
to the MEC-hosted global pool $\mathcal{M}$ together with a timestamp and UAV identifier (grow phase). 
The MEC server periodically aggregates all collected updates in batches (refine phase). 
During each aggregation, semantically similar prompts are identified via cosine similarity between their embeddings 
and merged using an EMA weighting scheme: 
the newer prompts slightly adjust the stored representation while preserving historical knowledge. 
This gradual fusion guarantees eventual consistency without requiring real-time synchronization 
and keeps communication overhead minimal. 
In this way, geographically dispersed UAVs can contribute updates asynchronously 
while the MEC node periodically harmonizes them into a unified, stable prompt memory.}

\blue{The quantitative evaluation of this conversion process is presented in Sec.~\ref{subsec:distill}.}

\subsection{AdaptFly Algorithm Overview}

Algorithm~\ref{alg:adaptfly} illustrates the unified AdaptFly framework for TTA across heterogeneous UAV agents. The procedure includes four main stages:

\begin{enumerate}[leftmargin=*, label=\textbf{Step \arabic*.}]
    \item Distribution Statistics (Lines 5-7):  
    For each UAV, stem $f^s_{\theta}$ of the backbone extracts low-level statistics from inputs. These are converted to activation features $\hat{\phi}^u_t$ and smoothed into moving averages $\phi^u_d$ using EMA to track domain characteristics over time.

    \item Token Prompt Retrieval for $\mathcal{U}_\ell$ (Lines 8-12): 
    If a domain shift is detected via divergence $D(\phi_d^L,\hat{\phi}_t^L) > z$, a query embedding $\mathbf{q}^L$ is generated. The system then retrieves and concatenates token prompts from memory $\mathcal{M}$ using cosine similarity, which are prepended to the patch sequence for inference.

    \item Sparse Prompt Optimization for $\mathcal{U}_h$ (Lines 13-20):  
    If domain shift is detected for $\mathcal{U}_h$, a sparse uncertainty-based mask $\mathbf{M}^H$ is generated, and CMA-ES is used to optimize a prompt $\mathbf{p}^H$ that minimizes prediction entropy. Otherwise, the previously optimized prompt is warped to the current frame using optical flow.

    \item Knowledge Consolidation (Lines 15-16):  
    When $\mathcal{U}_h$ completes optimization, the new sparse prompt and its domain embedding $\mathbf{q}^H$ are added to global memory $\mathcal{M}$, enabling future retrieval and transfer to resource-limited UAVs.
\end{enumerate}

\begin{algorithm}[t]
\DontPrintSemicolon
\caption{AdaptFly}
\label{alg:adaptfly}

\KwIn{%
Frozen backbone $f_{\theta}$ with stem $f^s_{\theta}$;\;
Test streams $\{x^H_t\}_{t=1}^T$ ($\mathcal{U}_h$) and $\{x^L_t\}_{t=1}^T$ ($\mathcal{U}_\ell$);\;
Global token prompt memory $\mathcal{M}$ (Eq.\,(6));\;
Sparsity ratio $\rho$; Fitness function $F$ (Eq.\,(9));\;
EMA factor $\lambda$, drift threshold $z$
}

\KwOut{Predictions $\{\hat{y}_t^H\},\ \{\hat{y}_t^L\}$}

\textbf{Init:}~$\phi_d^H\!\leftarrow\!0$, $\phi_d^L\!\leftarrow\!0$,\;
$\mathbf{p}^H_{\text{current}}\!\leftarrow\!\mathbf{0}$, $\mathbf{p}^H_{\text{prev}}\!\leftarrow\!\mathbf{0}$, $\mathbf{M}^H_{\text{current}}\!\leftarrow\!\mathbf{0}$\;
$\mathbf{P}_{\!\text{sel}}\!\leftarrow\!\varnothing$\;

\For{$t = 1,2,\dots,T$}{
    \For{$u\in\{H,L\}$}{
        $\hat{\phi}^u_t \gets \text{ComputeStats}(f^s_{\theta}(x^u_t))$\;
        $\phi^u_d \leftarrow \lambda\hat{\phi}^u_t + (1{-}\lambda)\phi^u_d$\;
    }

    \If{$D(\phi_d^L,\hat{\phi}_t^L) > z$}{
        $\mathbf{q}^L \leftarrow g(x_t^L)$\;
        $\mathcal{I} \leftarrow \textsc{TopN}(\mathbf{q}^L,\mathcal{M})$\;
        $\mathbf{P}_{\!\text{sel}} \leftarrow \texttt{Concat}(\{\mathbf{P}_j\}_{j\in\mathcal{I}})$\;
    }
    $\hat{y}_t^L \leftarrow f_{\theta}([\mathbf{P}_{\!\text{sel}};\;\mathbf{Z}_0(x_t^L)])$\;

    \If{$D(\phi_d^H,\hat{\phi}_t^H) > z$}{
        Invoke Algorithm~\ref{alg:svp_cma-es} for SVP optimization\;
        Convert $\mathbf{p}^H_{\text{current}}$ into a compatible token prompt $\mathbf{P}_{\text{new}}$ using Eq.~\eqref{eq:convert}\;
        Append $(\mathbf{q}^H, \mathbf{P}_{\text{new}})$ to the global prompt memory $\mathcal{M}$\;
    } \Else {
        $\mathbf{p}^H_{\text{current}} \leftarrow$ warp $\mathbf{p}^H_{\text{prev}}$ using optical flow between $x_{t-1}^H$ and $x_t^H$\;
    }
    $\hat{y}_t^H \leftarrow f_{\theta}(x_t^H + \mathbf{p}^H_{\text{current}})$\;
    $\mathbf{p}^H_{\text{prev}} \leftarrow \mathbf{p}^H_{\text{current}}$\;
}
\end{algorithm}

\begin{table}[!t]
  \centering
  \caption{Class definitions in UAVid and VDD datasets.}
  \label{tab:datasets}
  \vspace{0.5em}
  \begin{tabular}{c|l|l}
    \toprule
    \textbf{Dataset} & \textbf{Class ID} & \textbf{Class Name} \\\midrule
    \multirow{8}{*}{UAVid} 
        & 0 & Building \\
        & 1 & Road \\
        & 2 & Static Car \\
        & 3 & Tree \\
        & 4 & Low Vegetation \\
        & 5 & Human \\
        & 6 & Moving Car \\
        & 7 & Clutter \\\midrule
    \multirow{7}{*}{VDD} 
        & 0 & Others \\
        & 1 & Wall \\
        & 2 & Road \\
        & 3 & Vegetation \\
        & 4 & Vehicle \\
        & 5 & Roof \\
        & 6 & Water \\
    \bottomrule
  \end{tabular}
\end{table}

\begin{table*}[!t]
\centering
\small
\renewcommand{\arraystretch}{1.2}
\setlength{\tabcolsep}{4pt}
\caption{Comparison of TTA methods on the Cityscapes→VDD benchmark. Results are reported on four sub-domains: Urban (URB), Industrial (IND), Rural (RUR), and Natural (NAT). All values are averaged over three runs.}
\label{tab:vdd-tta}
\begin{tabular}{l|cc|cc|cc|cc|c}
\toprule
\multirow{2}{*}{\textbf{Method}} &
\multicolumn{2}{c|}{Source2URB} &
\multicolumn{2}{c|}{Source2IND} &
\multicolumn{2}{c|}{Source2RUR} &
\multicolumn{2}{c|}{Source2NAT} &
\multirow{2}{*}{Mean-mIoU} \\
& mIoU↑ & mAcc↑ & mIoU↑ & mAcc↑ & mIoU↑ & mAcc↑ & mIoU↑ & mAcc↑ & \\
\midrule
Source (NoAdapt)        & 66.2\tiny($\pm$0.3) & 80.3\tiny($\pm$0.2) & 63.1\tiny($\pm$0.4) & 77.5\tiny($\pm$0.3) & 57.7\tiny($\pm$0.5) & 73.9\tiny($\pm$0.4) & 54.3\tiny($\pm$0.6) & 71.1\tiny($\pm$0.3) & 60.3 \\
TENT                    & 65.8\tiny($\pm$0.3) & 80.0\tiny($\pm$0.3) & 62.7\tiny($\pm$0.3) & 77.1\tiny($\pm$0.3) & 57.4\tiny($\pm$0.4) & 73.6\tiny($\pm$0.4) & 53.9\tiny($\pm$0.5) & 70.7\tiny($\pm$0.4) & 60.0 \\
CoTTA                   & 66.8\tiny($\pm$0.3) & 81.0\tiny($\pm$0.3) & 64.0\tiny($\pm$0.3) & 78.1\tiny($\pm$0.3) & 58.5\tiny($\pm$0.3) & 75.1\tiny($\pm$0.3) & 55.0\tiny($\pm$0.4) & 71.8\tiny($\pm$0.4) & 61.1 \\
AgPT                    & 69.7\tiny($\pm$0.2) & 81.5\tiny($\pm$0.2) & 66.9\tiny($\pm$0.3) & 78.9\tiny($\pm$0.3) & 61.2\tiny($\pm$0.3) & 75.7\tiny($\pm$0.2) & 58.3\tiny($\pm$0.3) & 72.5\tiny($\pm$0.3) & 64.0 \\
SVDP                    & 71.3\tiny($\pm$0.2) & 82.4\tiny($\pm$0.2) & 69.0\tiny($\pm$0.2) & 80.0\tiny($\pm$0.2) & 63.4\tiny($\pm$0.3) & 76.9\tiny($\pm$0.2) & 60.6\tiny($\pm$0.3) & 74.1\tiny($\pm$0.3) & 66.1 \\
AdaptFly-Limited (Ours) & 70.0\tiny($\pm$0.2) & 82.2\tiny($\pm$0.2) & 67.8\tiny($\pm$0.3) & 79.5\tiny($\pm$0.2) & 62.0\tiny($\pm$0.2) & 76.3\tiny($\pm$0.2) & 60.0\tiny($\pm$0.3) & 73.5\tiny($\pm$0.3) & 65.0 \\
AdaptFly-Massive (Ours) & \textbf{72.1\tiny($\pm$0.1)} & \textbf{82.6\tiny($\pm$0.2)} & \textbf{70.3\tiny($\pm$0.2)} & \textbf{80.8\tiny($\pm$0.2)} & \textbf{64.7\tiny($\pm$0.2)} & \textbf{77.6\tiny($\pm$0.2)} & \textbf{62.9\tiny($\pm$0.2)} & \textbf{75.2\tiny($\pm$0.3)} & \textbf{67.5} \\
\bottomrule
\end{tabular}
\end{table*}

\section{Experiments}
\subsection{Experimental Setup}
\subsubsection{Datasets and Scenarios} To evaluate AdaptFly for UAV semantic segmentation, we use two public benchmarks: \textbf{UAVid}~\cite{lyu2020uavid} and \textbf{VDD}~\cite{cai2025vdd}. All experiments are conducted using an NVIDIA RTX 4090 GPU with 24GB memory under CUDA 11.5.

The UAVid dataset~\cite{lyu2020uavid} focuses on high-resolution urban scene understanding from oblique-view UAVs. It comprises 420 images extracted from 30 video sequences, captured at a typical UAV altitude of 50 meters. Each image has a resolution of either $4096 \times 2160$ or $3840 \times 2160$ pixels, offering rich visual content in complex city environments. UAVid presents substantial challenges including scale variation, motion blur, and frequent occlusions, reflecting real-world deployment conditions. The dataset defines 8 semantic classes: \textit{building}, \textit{road}, \textit{static car}, \textit{tree}, \textit{low vegetation}, \textit{human}, \textit{moving car}, and \textit{background clutter}, covering both static and dynamic elements.

Due to large image sizes, a multi-stage preprocessing strategy is adopted. During training and validation, a clipping-based approach is applied using fixed $512 \times 512$ patches with a stride of 256 pixels, resulting in approximately 8000 training and 2800 validation samples. In contrast, test-time inference preserves full image resolution and employs a sliding window approach with a window size of $1024$ pixels and an overlap of $128$ pixels between adjacent patches. 

The VDD dataset~\cite{cai2025vdd} broadens the UAV segmentation landscape by including rural, industrial, and diverse low-altitude environments. All images are captured at varied flight altitudes and perspectives, standardized to a resolution of $2048 \times 1024$ pixels. Compared to UAVid, VDD introduces more significant domain shifts across lighting conditions, terrains, and object appearances. It covers 7 broad semantic categories, enabling evaluation of adaptation methods under diverse spatial and environmental variations. Detailed class definitions for both UAVid and VDD datasets are summarized in Table~\ref{tab:datasets}.

\subsubsection{Baselines and Implementation Details} We compare AdaptFly against the following baselines, covering both conventional and prompt-based TTA strategies:
(1) NoAdapt: A non-adaptive baseline where the source-pretrained segmentation model is directly applied to target domains without any modification.
(2) TENT~\cite{wangtent}: A TTA method that minimizes prediction entropy by updating batch normalization affine parameters on unlabeled target data.
(3) CoTTA~\cite{wang2022continual}: A continual TTA framework that mitigates error accumulation via teacher-student consistency and stochastically restores parts of the model to avoid catastrophic forgetting.
(4) SVDP~\cite{yang2024exploring}: An SVP learning approach that allocates trainable tokens to uncertainty-aware spatial locations for dense prediction tasks under domain shift.
(5) Attention-guided prompt tuning (AgPT)~\cite{gao2024attention}: An attention-guided prompt tuning method that dynamically modulates the influence of prompts across layers to adapt vision foundation models without access to source data.
(6) AdaptFly-Limited/Massive: Our proposed lightweight (token-retrieval) and powerful (CMA-ES optimized) variants of prompt-guided TTA tailored for heterogeneous UAV agents.

We use three evaluation metrics to comprehensively assess adaptation performance:
(1) Mean Class Accuracy (mAcc\%): average of per-class accuracies, where each class accuracy is defined as the ratio of correctly predicted pixels to the total number of ground-truth pixels for that class. This metric reflects class-wise recall and provides insight into performance balance across all categories, especially for rare classes.
(2) Class-wise Intersection over Union (IoU\%): the IoU score computed independently for each semantic class, measuring overlap between predicted and ground-truth masks. It highlights how well each class is localized and segmented.
(3) Mean Intersection over Union (mIoU\%): the average of class-wise IoUs, serving as the primary benchmark for semantic segmentation performance. It balances both false positives and false negatives and is widely adopted in UAV segmentation benchmarks such as UAVid and VDD.

\subsection{Cross Scenarios Adaptation Evaluation}
\subsubsection{Cityscapes-to-VDD Benchmark}
To characterize distribution shifts introduced by diverse environments in the Cityscapes→VDD cross-domain setting, we partition the VDD dataset into four mutually exclusive sub-domains: Urban (URB), Industrial (IND), Rural (RUR), and Natural (NAT). They are defined according to typical landform and functional attributes. These sub-domains exhibit pronounced differences in visual appearance and object composition, thereby providing representative domain shifts for evaluation. The segmentation model is pre-trained on the Cityscapes dataset and then tested or adapted on each VDD sub-domain.

Table~\ref{tab:vdd-tta} presents a quantitative comparison of TTA methods on the Cityscapes$\rightarrow$VDD setting. The results highlight significant performance differences across sub-domains, confirming challenges of generalizing segmentation models to diverse target environments.

The NoAdapt baseline achieves 60.3 mIoU on average, indicating severe performance degradation without adaptation. Classical TTA methods such as TENT and CoTTA yield limited improvements, with mean-mIoU scores of 60.0 and 61.1, respectively. Although CoTTA performs marginally better in urban and rural scenes, its effectiveness diminishes in natural domains (e.g., 55.0 mIoU on NAT).

Prompt-based methods demonstrate stronger generalization under shift. AgPT achieves 63.9 mIoU by injecting attention-weighted prompts, while SVDP further improves to 66.1 mIoU using uncertainty-aware sparse prompt learning.

Our proposed AdaptFly consistently outperforms prior baselines in both limited and massive resource settings. AdaptFly-Limited achieves 65.0 mIoU using a retrieval-only, training-free strategy with minimal computation. AdaptFly-Massive leverages CMA-ES optimization and achieves the best results across all sub-domains, with a mean-mIoU of 67.5. Notably, it delivers substantial gains in challenging natural scenes (e.g., +8.6 mIoU over Source on NAT).

\subsubsection{Cityscapes-to-UAVid Benchmark}

Table~\ref{tab:uavid-tta} reports class-wise IoU and overall mIoU for different TTA methods on the Cityscapes$\rightarrow$UAVid setting. The baseline model (NoAdapt), trained solely on Cityscapes, yields 66.2 mIoU and suffers from particularly poor performance on challenging categories such as \textit{Human} (29.0) and \textit{Static car} (58.4), highlighting its limited generalization to oblique-view UAV imagery.

Traditional TTA methods such as TENT and CoTTA achieve only marginal gains, reaching 67.0 and 69.2 mIoU, respectively. Although CoTTA improves on dynamic categories like \textit{Moving car} (73.4), its effectiveness on rare or occluded classes remains limited.

Prompt-based methods exhibit stronger performance. AgPT and SVDP achieve 69.8 and 70.5 mIoU, with notable improvements on categories like \textit{Tree} and \textit{Road}. Our proposed AdaptFly-Limited method, despite requiring no test-time training, surpasses all baselines with 70.2 mIoU, indicating the strength of retrieval-based prompt adaptation.

Notably, AdaptFly-Massive achieves the best results across nearly all categories, including significant improvements on \textit{Human} (36.2) and \textit{Static car} (69.4), leading to a new state-of-the-art mIoU of 73.0. These results validate that AdaptFly can effectively address domain shifts between Cityscapes and UAVid, especially under challenging conditions such as occlusion, small objects, and dense urban scenes.

\begin{table*}[!t]
\centering
\small
\renewcommand{\arraystretch}{1.15}
\setlength{\tabcolsep}{5pt}
\caption{Class-wise IoU (\%) and overall mIoU on the \textbf{UAVid} dataset.  
The results are averaged over three runs; standard deviation is shown in parentheses.  
The best score for each column is \textbf{bold}.}
\label{tab:uavid-tta}
\begin{tabular}{l|cccccccc|c}
\toprule
\textbf{Method} &
Clutter & Buildings & Road & Tree & Low veg. &
Moving car & Static car & Human & mIoU \\
\midrule
Source (NoAdapt)        & 65.6\tiny($\pm$0.3) & 86.0\tiny($\pm$0.2) & 78.3\tiny($\pm$0.4) & 79.3\tiny($\pm$0.3) & 63.0\tiny($\pm$0.3) & 70.0\tiny($\pm$0.4) & 58.4\tiny($\pm$0.3) & 29.0\tiny($\pm$0.2) & 66.2 \\
TENT                    & 66.4\tiny($\pm$0.3) & 86.6\tiny($\pm$0.2) & 79.2\tiny($\pm$0.3) & 79.8\tiny($\pm$0.2) & 63.5\tiny($\pm$0.2) & 71.3\tiny($\pm$0.3) & 58.7\tiny($\pm$0.3) & 30.0\tiny($\pm$0.2) & 67.0 \\
CoTTA                   & 68.9\tiny($\pm$0.2) & 88.5\tiny($\pm$0.2) & 80.2\tiny($\pm$0.3) & 80.5\tiny($\pm$0.3) & 65.2\tiny($\pm$0.3) & 73.4\tiny($\pm$0.3) & 65.0\tiny($\pm$0.2) & 32.2\tiny($\pm$0.2) & 69.2 \\
AgPT                    & 69.5\tiny($\pm$0.2) & 88.8\tiny($\pm$0.2) & 80.8\tiny($\pm$0.3) & 81.0\tiny($\pm$0.3) & 65.9\tiny($\pm$0.3) & 73.8\tiny($\pm$0.2) & 65.9\tiny($\pm$0.2) & 32.8\tiny($\pm$0.2) & 69.8 \\
SVDP                    & 70.3\tiny($\pm$0.2) & 89.0\tiny($\pm$0.2) & 82.0\tiny($\pm$0.2) & 81.2\tiny($\pm$0.2) & 66.0\tiny($\pm$0.2) & 76.2\tiny($\pm$0.2) & 66.2\tiny($\pm$0.2) & 32.8\tiny($\pm$0.2) & 70.5 \\
AdaptFly-Limited (Ours) & 70.2\tiny($\pm$0.2) & 88.4\tiny($\pm$0.2) & 82.5\tiny($\pm$0.2) & 80.8\tiny($\pm$0.2) & 64.5\tiny($\pm$0.2) & 75.6\tiny($\pm$0.2) & 66.3\tiny($\pm$0.2) & 32.6\tiny($\pm$0.3) & 70.2 \\
AdaptFly-Massive (Ours) & \textbf{72.1}\tiny($\pm$0.2) & \textbf{89.2}\tiny($\pm$0.2) & \textbf{84.8}\tiny($\pm$0.2) & \textbf{83.6}\tiny($\pm$0.2) & \textbf{68.3}\tiny($\pm$0.2) & \textbf{79.4}\tiny($\pm$0.2) & \textbf{69.4}\tiny($\pm$0.2) & \textbf{36.2}\tiny($\pm$0.2) & \textbf{73.0} \\
\bottomrule
\end{tabular}
\end{table*}

\begin{figure*}[t]
    \centering
    \includegraphics[width=1\linewidth]{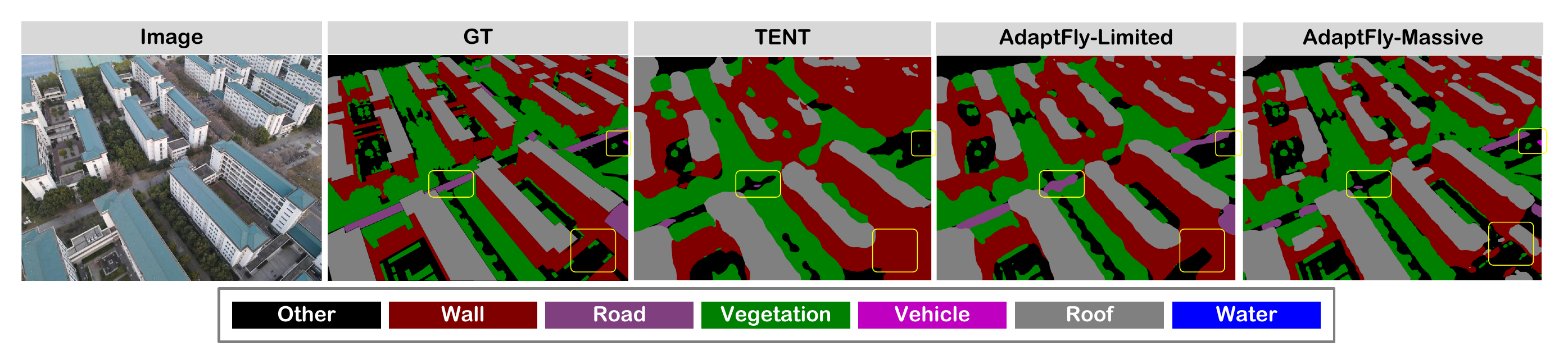}
    \caption{Example of semantic segmentation results on the VDD dataset. GT denotes the ground truth annotations. Distinct differences in segmentation accuracy among methods (TENT, AdaptFly-Limited, and AdaptFly-Massive) are highlighted with yellow boxes.}
    \label{fig:vdd_infer}
    \vspace{-0.1in}
\end{figure*}

\begin{table*}[!t]
  \centering
  \small
  \renewcommand{\arraystretch}{1.15}
  \setlength{\tabcolsep}{5pt}
  \caption{Latency-mIoU trade-off of AdaptFly variants  
           (batch size = 1, profiled on RTX\,4090).}
  \label{tab:adaptfly-speed}
  \begin{tabular}{lccc @{\hskip 8pt} ccc @{\hskip 8pt} ccc}
    \toprule
    \multirow{2}{*}{\textbf{Model}} &
    \multirow{2}{*}{\textbf{Params (M)}} &
    \multirow{2}{*}{\textbf{Input}} &
    \multicolumn{3}{c}{\textbf{AdaptFly-Limited}} &
    \multicolumn{3}{c}{\textbf{AdaptFly-Massive}} \\
    \cmidrule(lr){4-6} \cmidrule(lr){7-9}
    & & \textbf{size} & mIoU (\%) & Latency (ms) & FPS & mIoU (\%) & Latency (ms) & FPS \\
    \midrule
    SegFormer-B0 & 15  & $512\times512$ & 66.2 &  7.0  & 142.9 & 68.5 & 15.5  &  64.5 \\
    SegFormer-B3 & 189 & $512\times512$ & 70.2 & 40.0  &  25.0 & 73.0 & 80.4  &  12.4 \\
    SegFormer-B5 & 339 & $512\times512$ & 71.0 & 80.0  &  12.5 & 73.9 & 205.1 &  4.9 \\
    \bottomrule
  \end{tabular}
\end{table*}

\subsection{Quantitative and Qualitative Analysis}
\subsubsection{Quantitative Analysis}
To evaluate runtime performance of AdaptFly, we measure latency and frames per second (FPS) on three SegFormer variants with an input resolution of $512{\times}512$ and batch size of 1. As shown in Table~\ref{tab:adaptfly-speed}, SegFormer-B0 achieves a favorable trade-off between accuracy and efficiency: it delivers 66.2\% mIoU under the AdaptFly-Limited setting with only 7.0 ms latency and 142.9 FPS. Under the AdaptFly-Massive setting, it still maintains high throughput (64.5 FPS) while boosting accuracy to 68.5\%.

In comparison, \textbf{SegFormer-B3} offers improved segmentation performance (up to 73.0\% mIoU) at the cost of higher latency (80.4 ms) and reduced FPS (12.4) when used with AdaptFly-Massive. Finally, the largest model SegFormer-B5 achieves the best accuracy (73.9\% mIoU) but incurs substantial computational overhead with a latency of 205.1 ms and FPS of only 4.9.

These results suggest that SegFormer-B0 with AdaptFly-Limited is well-suited for real-time, resource-limited UAV agents, while AdaptFly-Massive with B3/B5 is more appropriate for high-end edge devices prioritizing segmentation quality over speed.

\begin{figure*}[t]
    \centering
    \includegraphics[width=1\linewidth]{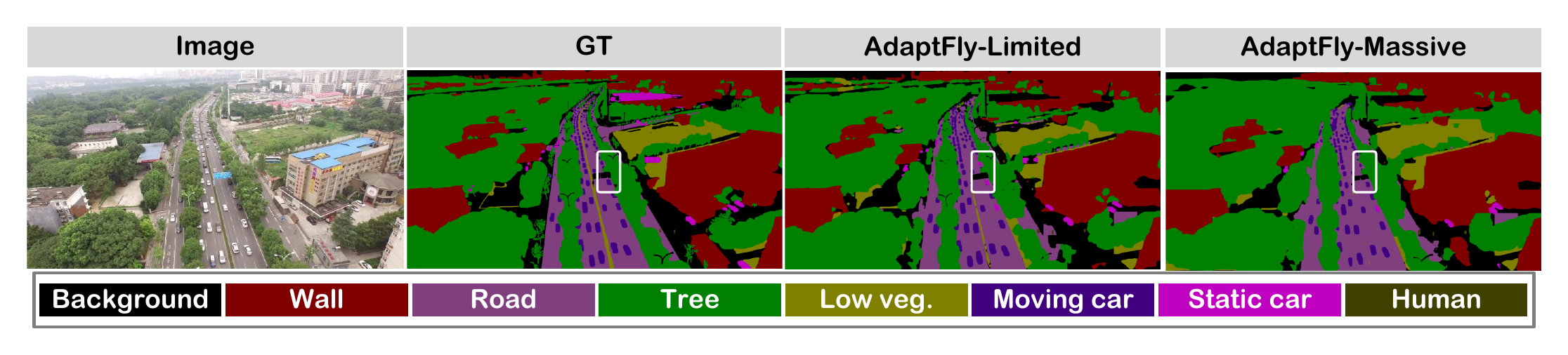}
    \caption{Example of semantic segmentation results on the UAVid dataset. Distinct differences in segmentation accuracy among methods (AdaptFly-Limited, and AdaptFly-Massive) are highlighted with white boxes.}
    \label{fig:uavid_infer}
    \vspace{-0.1in}
\end{figure*}

\subsubsection{Qualitative Analysis on VDD and UAVid Datasets}

Fig.~\ref{fig:vdd_infer} presents a representative sample from the VDD dataset, including the input aerial image, corresponding ground truth mask, and semantic segmentation results produced by TENT, AdaptFly-Limited, and AdaptFly-Massive. All three methods demonstrate strong performance in major classes such as building, roof, and vegetation.

However, for the human category, both TENT and AdaptFly-Limited struggle with accurate prediction, while AdaptFly-Massive shows a slight improvement. Interestingly, for the road category, AdaptFly-Limited outperforms AdaptFly-Massive, indicating that lightweight models can sometimes offer better generalization in specific classes under resource-limited scenarios.
These results highlight complementary strengths of the two AdaptFly variants under different scene complexities and category distributions.

Fig.~\ref{fig:uavid_infer} presents visual results on the UAVid dataset, including original aerial images, corresponding ground-truth masks, and segmentation predictions produced by AdaptFly. A closer inspection of specific regions reveals that AdaptFly-Limited sometimes struggles to distinguish between static and moving vehicles. This ambiguity is largely attributed to visual similarity and limited contextual cues of stationary vehicles, which often resemble backgrounds or other classes in low-altitude scenes.

In contrast, AdaptFly-Massive demonstrates improved discrimination between static and dynamic vehicles. This advantage arises from the use of SVP, which guide the model attention toward high-uncertainty regions. As illustrated in Fig.~\ref{fig:combined-flow}(a), the optimized prompt $\mathbf{p}^{\star}$ is propagated to subsequent frames through a lightweight optical-flow-based warping mechanism (``Warp") and refreshed only when significant changes in the uncertainty map are detected. 

\subsection{Real-World Deployment}

\begin{figure}[t]
    \centering
    \includegraphics[width=0.9\linewidth]{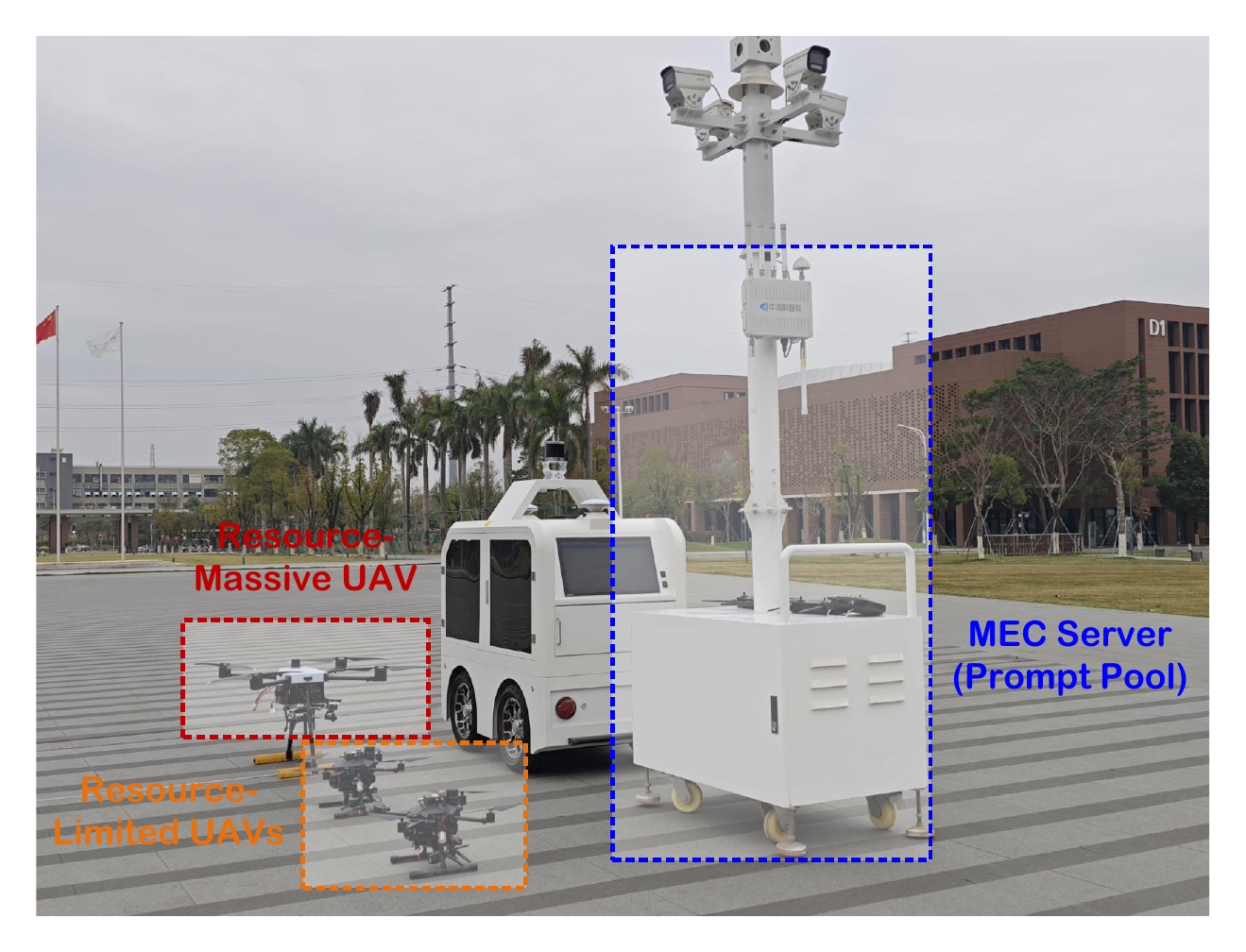}
    \caption{The testbed with UAVs and MEC server.}
    \label{fig:deployment-setup}
    \vspace{-0.1in}
\end{figure}

To validate practical feasibility and robustness of the AdaptFly framework in real UAV systems, we construct a real-world deployment platform consisting of heterogeneous UAVs and edge devices, as illustrated in Fig.~\ref{fig:deployment-setup}. The system includes three key components. 

First, a resource-massive UAV is deployed with sufficient onboard computational capacity to run the full AdaptFly-Massive pipeline. It supports real-time SVP optimization via CMA-ES and simulates a high-end agent capable of local domain adaptation. Once optimized, visual prompts are uploaded to the global prompt pool for shared use. 

Second, two resource-limited UAVs are deployed as lightweight aerial agents with constrained hardware incapable of performing prompt optimization. These UAVs operate under the AdaptFly-limited mode, retrieving relevant prompts from memory through embedding-based similarity search. This forward-only adaptation strategy enables training-free, low-latency semantic segmentation in unseen environments. 

Third, a MEC server acts as the shared memory host and communication hub. Deployed on a ground vehicle and equipped with 5G connectivity, it maintains the global prompt pool $\mathcal{M}$ and handles dynamic key-value prompt updates from massive UAVs. In turn, it supports online query responses from limited UAVs, thereby enabling cross-device knowledge sharing and continual refinement of domain expertise.

The experiments are conducted in a campus-scale outdoor area featuring diverse semantic categories (e.g., buildings, roads, vegetation). As shown in Fig.~\ref{fig:real_compare}(a)(b), we evaluate the system under various weather conditions, including overcast and heavy rain scenarios, to assess its adaptability in real dynamic environments.

\begin{figure}[t]
    \centering
    \includegraphics[width=1\linewidth]{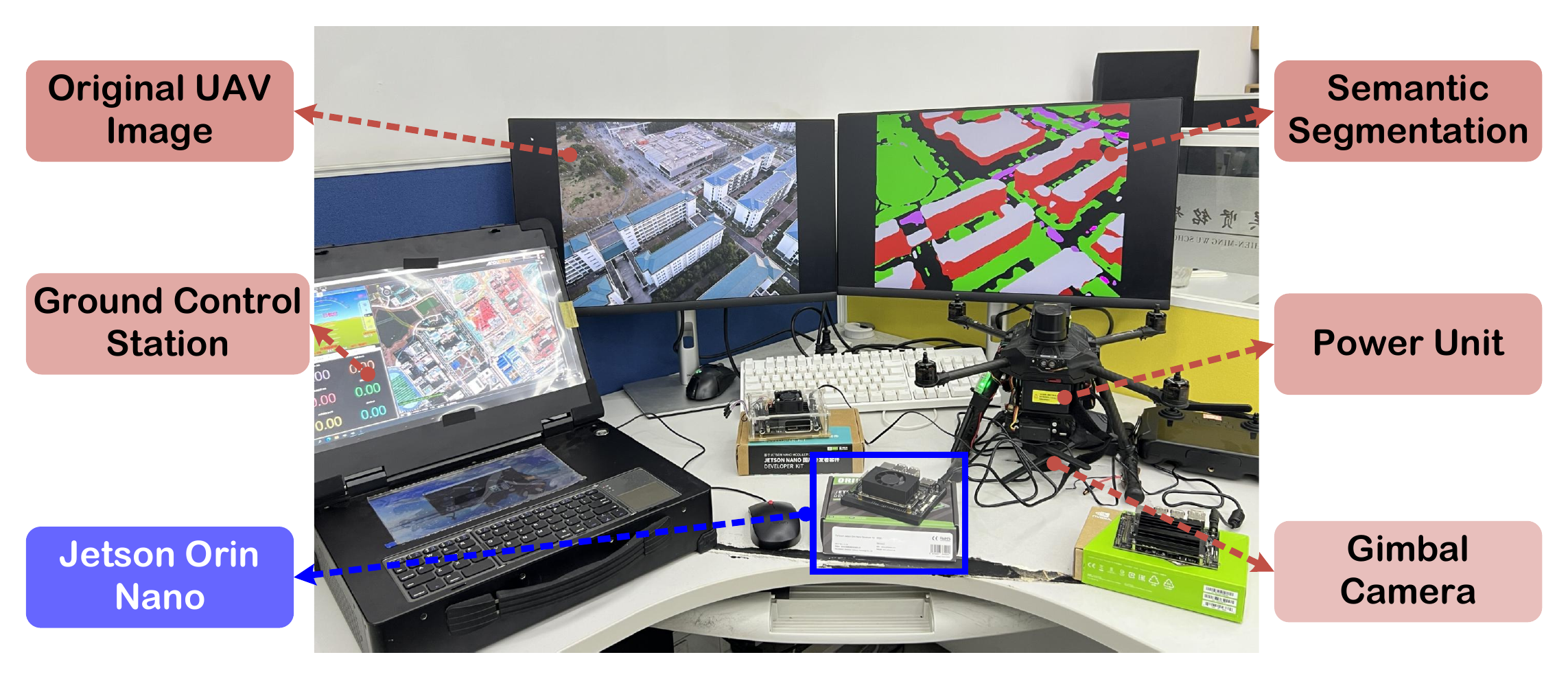}
    \caption{UAV and edge device configuration with the ground control station for mission planning and control.}
    \label{fig:Edge_Device}
    \vspace{-0.1in}
\end{figure}

\subsubsection{Qualitative Comparison Under Weather Shifts}

\begin{figure*}[t]
    \centering
    \includegraphics[width=0.8\linewidth]{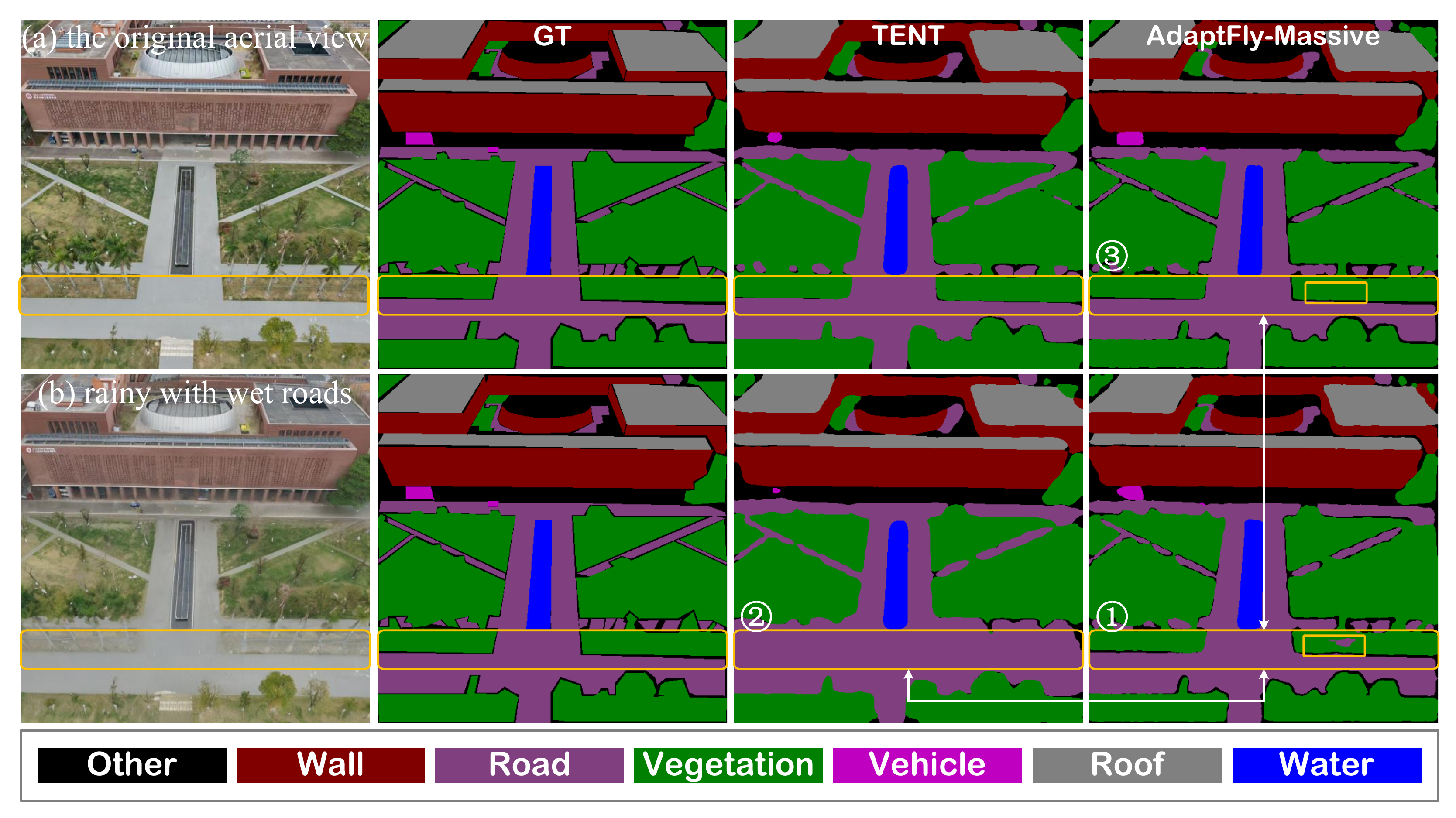}
    \caption{Qualitative comparison under weather-induced domain shifts. Distinct differences in segmentation accuracy among methods (TENT and AdaptFly-Massive) are highlighted with yellow boxes.}
    \label{fig:real_compare}
    \vspace{-0.1in}
\end{figure*}

Fig.~\ref{fig:real_compare} presents a visual comparison of segmentation performance under two weather conditions in the same UAV-captured scene. Subfigure (a) depicts the environment under normal visibility, while subfigure (b) reflects the same location under rainy conditions, where road surfaces become darker and more reflective, posing a challenge for visual perception.

In the clear-weather setting (a), both TENT and AdaptFly-Massive perform comparably well, correctly identifying most semantic categories with minimal error. However, under rainy conditions (b), a performance gap emerges. The yellow rectangle highlights a low-visibility region where water accumulation and illumination changes obscure semantic boundaries.

Comparing annotations \ding{172} and \ding{174} (AdaptFly under different conditions), we observe that while minor misclassifications are introduced under rain, AdaptFly maintains stable predictions, with accurate delineation of roads and vegetation. In contrast, comparison between \ding{172} and \ding{173} shows that TENT suffers from a significant drop in segmentation quality in (b), failing to preserve class boundaries and producing noisy predictions.

These results validate effectiveness of AdaptFly-Massive in maintaining robustness under domain shifts caused by environmental variations. By leveraging test-time prompt optimization, AdaptFly adapts more gracefully to weather-induced appearance changes, outperforming baseline TTA methods.

\subsubsection{Edge Profiling on Resource-Massive UAV Hardware}
To validate the practical feasibility of $\mathcal{U}_h$ deployment, we further profile the CMA–ES optimization on an embedded edge-AI platform representative of high-end onboard computing. 
We use the NVIDIA Jetson Orin Nano 8 GB (67 TOPS, 1024 Ampere Tensor Cores, 1020 MHz) configured with TensorRT FP16 inference. 
The same CMA–ES settings as in Sec.~V-B are applied (population = 8, generations = 5).
Each optimization is performed on \(512\times512\) images using SegFormer-B0 as the frozen backbone. 
The average latency and power are measured with \texttt{tegrastats}.

\begin{table*}[t]
\centering
\caption{Profiling CMA–ES adaptation on realistic edge hardware.}
\label{tab:edge_profiling}
\begin{tabular}{lcccc}
\toprule
Device & Power (W) & Adapt Time (s/img) & $\Delta$mIoU (\%) & Note \\
\midrule
RTX 4090 & 450 & 0.15 & +4.5 & desktop GPU \\
Jetson Orin Nano 8 GB & 15 & 1.10 & +3.8 & onboard edge GPU \\
Jetson Orin Nano 4 GB & 10 & 1.80 & +3.6 & low-power variant \\
\bottomrule
\end{tabular}
\end{table*}

\blue{Results in Table~\ref{tab:edge_profiling} show that CMA–ES optimization remains feasible on the Orin Nano platform, 
achieving full adaptation within $\approx$ 1 to 2 s per image at 10 to 15 W power draw. 
These findings confirm that our definition of resource-massive UAVs ($\geq$ 30 TOPS) matches current embedded hardware, 
and that AdaptFly’s optimization can be executed periodically rather than frame-by-frame to balance latency and energy consumption.}

\subsection{SVP-to-Token Distillation Evaluation}
\label{subsec:distill}

To quantitatively validate the SVP-to-token distillation process described in Sec.~\ref{subsec:convert}, we conduct an evaluation on the Cityscapes$\rightarrow$UAVid benchmark using SegFormer-B3.

\blue{We implement Eq.~\eqref{eq:convert} as a lightweight regression on the MEC server \emph{without} updating backbone parameters.
Let $f_\theta^{\text{enc}}(\cdot)$ be the last-stage encoder tokens (after LayerNorm, flattened).
The \emph{teacher} representation is $T = f_\theta^{\text{enc}}(x_t^{H}+p^{H})$ (no gradient);
the \emph{student} representation is $S = f_\theta^{\text{enc}}([P;Z_0(x_t^{H})])$.
We optimize only $P\in\mathbb{R}^{L\times d}$ (by default $L{=}8$ tokens, $d{=}768$) using Adam with learning rate $1\!\times\!10^{-4}$, batch size $1$, for a small, fixed number of steps.
Unless otherwise stated, we use $8$ gradient steps and FP16 storage for the final $P_{\text{new}}$.
This produces a compact prompt entry of $\approx 24$--$32$\,KB (FP16/FP32), which is then inserted into the global memory $\mathcal{M}$ with its domain key.}

\textbf{MEC setup and protocol.}
All distillation runs are executed \textit{off the critical path} on a mobile MEC server equipped with a single RTX~4090 (24\,GB).
A distillation job is triggered once drift is detected and $\mathcal{U}_h$ has finished CMA–ES.
For amortization, we use $K{=}5$ unlabeled frames from the same domain window (e.g., consecutive or near-by frames); the loss in \eqref{eq:convert} is averaged over these frames.
The resulting $P_{\text{new}}$ is pushed to $\mathcal{M}$ together with the domain embedding $q^{H}$ so that future $\mathcal{U}_\ell$ can retrieve it with a single forward pass.

\subsubsection{Effectiveness}
Table~\ref{tab:distill_main} quantifies the effectiveness on the Cityscapes$\rightarrow$UAVid setting with SegFormer-B3.
Compared to a pre-trained retrieval-only token prompt, the distilled tokens recover most of the gain achieved by SVP while requiring only a one-time offline cost on the MEC server. 

\blue{
These results show that token prompts distilled from the SVP of $\mathcal{U}_h$ recover 72.6 mIoU compared with 73.0 mIoU for the original SVP, while simple retrieval-only prompts achieve 70.2 mIoU.
Hence, sharing distilled prompts via the global pool improves $\mathcal{U}_\ell$ performance by about +2.4 mIoU without any additional onboard optimization.
}

\subsubsection{Distillation Cost} The average runtime is $280{\pm}30$ ms per prompt (averaged over three domains), which is a one-time offline cost.
After this step, the resulting token prompts can be reused by multiple $\mathcal{U}_\ell$ agents without any additional computation.
This confirms that the proposed distillation is lightweight and feasible for deployment on MEC-class hardware.

\begin{table}[t]
\centering
\caption{Effectiveness and cost of SVP-to-token distillation (Cityscapes$\rightarrow$UAVid, SegFormer-B3).
``Time'' reports the one-time offline wall-clock per distilled prompt on the RTX~4090 MEC server; retrieval has zero distillation time.}
\label{tab:distill_main}
\begin{tabular}{lccc}
\toprule
Method & Type & mIoU (\%) & Time (ms) \\
\midrule
SVP (original, $\mathcal{U}_h$) & optimized & \textbf{73.0} & -- \\
Token prompt (retrieved) & pre-trained & 70.2 & 0 \\
Distilled token (ours) & from SVP & 72.6 & $280 \pm 30$ \\
\bottomrule
\end{tabular}
\end{table}

\subsubsection{Convergence vs.\ Steps}
\blue{We further ablate the number of optimization steps (averaged over three domains).
Results in Table~\ref{tab:distill_steps} show that $5$-$8$ steps are sufficient; $10$ steps bring negligible improvements while increasing latency.}

\begin{table}[t]
\centering
\caption{Ablation on optimization steps (Cityscapes$\rightarrow$UAVid, SegFormer-B3).
Time is the one-time distillation cost on RTX~4090.}
\label{tab:distill_steps}
\begin{tabular}{cccc}
\toprule
Steps & 3 & 5 & 8 \;\; / \;\; 10 \\
\midrule
mIoU (\%) & 72.0 & 72.4 & \textbf{72.6} \;\; / \;\; 72.6 \\
Time (ms) & 120 & 190 & \textbf{280} \;\; / \;\; 350 \\
\bottomrule
\end{tabular}
\end{table}

\section{Conclusion}
We propose AdaptFly, a unified framework for heterogeneous UAV networks that integrates prompt-based tuning with TTA to enable collaborative perception across distributed UAV agents.
AdaptFly achieves training-free generalization of segmentation foundation models by combining lightweight prompt retrieval for resource-limited UAVs and gradient-free prompt optimization for resource-massive UAVs.
It adapts dynamically to domain shifts without modifying model weights.
Quantitative evaluations on UAVid and VDD datasets show clear gains over existing TTA methods in both segmentation accuracy and robustness. Qualitative results further confirm its effectiveness in challenging visual conditions such as occlusion and weather changes. 
Deployment on a real-world UAV testbed validates its practical applicability, showcasing stable performance in dynamic scenes. 
By enabling asynchronous knowledge sharing with negligible bandwidth, AdaptFly demonstrates how networked UAVs can collectively 
improve perception resilience. 
This resilience is a critical requirement for scalable low-altitude operations.
These findings suggest that AdaptFly provides essential support for achieving deployable, general-purpose perception across low-altitude UAV networks, contributing to more scalable and efficient sensing systems in the emerging low-altitude economy.

\bibliographystyle{IEEEtran}  
\bibliography{references}
\end{document}